# Convex-Neural RRT*: Fast and Reliable Learning-Guided Sampling for High-Quality Robot Path Planning

Hichem Cheriet, Badra Khellat Kihel, Samira Chouraqui, and Bara J. Emran

**Abstract**

Sampling-based algorithms for robot path planning offer probabilistic completeness and strong empirical convergence properties across environments with diverse obstacle configurations. However, in practice, these methods often require many iterations to obtain high-quality solutions. This paper proposes Convex-Neural RRT*, an enhanced RRT* variant that incorporates neural guidance to predict informative waypoint regions near high-quality paths. Convex candidate regions are extracted from these predictions, enabling the planner to concentrate exploration on geometrically relevant areas while preserving global exploration. The proposed algorithm is evaluated against Neural RRT*, Neural Informed RRT*, classical RRT*, and LTA* across three environment types and 18 benchmark maps. Experimental results show that Convex-Neural RRT* reduces computation time by 30–75% compared to neural-guided variants and up to 88–98% relative to LTA*, while achieving an average path length reduction of approximately 5% compared to classical RRT*, with larger improvements observed in complex environments. The method also maintains an overall success rate above 99% across varying obstacle densities. These findings indicate that convex-guided neural sampling provides an effective balance between computational efficiency and solution quality, supporting its applicability to time-sensitive robotic navigation tasks.

 

## 1 Introduction

Motion planning is a fundamental task in robotics, where a feasible, collision-free path must be computed for safe and efficient navigation. Planners are often required to handle environments of varying complexity, including convex and concave obstacles, static or dynamic obstacles, and known or unknown spaces. A high-quality path is typically characterized by its short length, smoothness, and minimal computation time.

Several families of planners have been proposed to address these challenges, each with distinct advantages and limitations. Artificial Potential Field (APF) methods (Khatib 1986) model the environment as a magnetic field, where free regions attract the robot while occupied regions repel it. These approaches are computationally light but often fail in narrow passages due to local minima, making them unreliable in cluttered environments.

To overcome such limitations, grid-based search methods such as A* (Hart et al. 1968), D* (Stentz 2003), and D* Lite (Koenig and Likhachev 2005) calculate optimal paths on discretized maps. However, their computational and memory costs increase significantly with environment dimensionality, which makes them less practical for 3D applications such as UAVs or manipulator motion planning.

Sampling-based planners, including Probabilistic Roadmaps (PRM) (Kavraki et al. 1996) and Rapidly-Exploring Random Trees (RRT, RRT*) (LAVALLE 1998; Karaman and Frazzoli 2011), construct paths by randomly sampling the configuration space until a feasible solution is found. These planners are effective in high-dimensional spaces but often require a large number of iterations to achieve near-optimal paths, and tuning parameters for each environment can be computationally intensive.

Optimization-based planners, such as Particle Swarm Optimization (PSO) (Kennedy and Eberhart 1995), Ant Colony Optimization (ACO) (Dorigo and Caro 1999), and Grey Wolf Optimizer (GWO) (Mirjalili et al. 2014), treat motion planning as a continuous optimization problem. These methods generate smooth, dynamically consistent trajectories but are sensitive to initialization and prone to getting trapped in local minima, especially in environments with narrow or non-convex passages.

Tangent or visibility-based planners construct paths by connecting critical points such as obstacle vertices, corners, or tangents, effectively exploiting the geometry of the environment. Examples include Visibility Graph (VG) (Lozano-Pérez and Wesley 1979; Liu and Arimoto 1992) and Tangent Bug Algorithm (Kamon et al. 1998). These planners are capable of finding the shortest paths in polygonal environments and provide a geometric understanding of obstacle-free regions. However, they are limited in handling complex 3D environments or highly irregular, concave obstacles, as the number of critical points can grow significantly, increasing computational cost.

These limitations motivate the integration of learning-based guidance with classical sampling and optimization methods to improve efficiency and reliability in cluttered environments. In this work, we propose Convex-Neural RRT*, a structured learning-guided sampling-based planner that combines neural guidance with convex-region exploitation. The main contributions of this paper are:

**Neural guidance for informative regions:** A U-Net model predicts regions near high-quality path waypoints, enabling focused sampling in geometrically relevant areas of the search space.

**Convex-guided hybrid exploration:** Convex corners are extracted from the predicted regions to guide RRT* expansion. A hybrid sampling strategy balances exploration within local convex hulls and global convex-corner waypoints, maintaining robustness even under imperfect predictions.

**Early stopping mechanism:** An adaptive termination criterion based on cost stabilization reduces unnecessary iterations and improves computational efficiency.

**Comprehensive experimental validation:** The algorithm is evaluated across multiple difficulty levels and benchmark maps, demonstrating consistent improvements in computational efficiency while maintaining competitive path quality, smoothness, and high success rates.

# 2 Related Works

In recent years, the path planning problem has seen significant advances powered by techniques that hybridize classical planners, introduce new variants, or integrate learning-based models. These developments aim to overcome the limitations of traditional methods while improving efficiency, optimality, and robustness.

Sampling techniques, which traditionally suffer from path sub-optimality and the need for large iteration counts, have seen important improvements. The RRT* algorithm adopts a rewiring technique after a solution is found to gradually improve the path toward optimality. To further reduce the search-space complexity and dimensionality, the Informed RRT* algorithm (Gammell et al. 2014) was introduced. It restricts sampling to an elliptical region that contains both the start and goal positions once an initial feasible solution is found. This approach has demonstrated strong performance across diverse scenarios. To handle dynamic environments where obstacles can change positions unpredictably, algorithms such as RT-RRT* (Naderi et al. 2015) and RRTX (Otte and Frazzoli 2014) were proposed. These variants maintain the RRT* structure while continuously rewiring the tree in response to environmental updates. RH-RRT* (Lee et al. 2017) is another recent variant specifically designed for unknown environments, where no prior information about obstacles is available. These algorithms have shown notable efficiency and practicality; however, they still fundamentally rely on randomized sampling, which cannot guarantee optimal solutions. They also tend to struggle in narrow passages and often require substantial computational resources depending on input parameters such as step size, iteration limits, and rewiring radius. Beyond RRT-based planners, other sampling approaches have also been proposed to improve efficiency. PRM-based methods such as PRM* (Karaman and Frazzoli 2011) and Lazy-PRM (Bohlin and Kavraki 2000) build global roadmaps that capture free space connectivity but still require many collision checks in cluttered maps. Batch-sampling planners like FMT* (Janson et al. 2015) and BIT* (Gammell et al. 2014) further reduce unnecessary exploration by expanding samples in ordered batches, showing faster convergence in many scenarios. However, these methods still depend on randomized sampling, which remains inefficient in narrow passages and complex non-convex environments where many samples are uninformative.

Recent graph-based planners focus on efficiency in 3D and complex environments. Szczerba et al. (Szczerba et al. 2000), and Li et al. (Li et al. 2002) proposed Sparse A* Search (SAS) for fast collision-free path computation for UAVs. Liu et al. (Liu et al. 2017) developed a 3D Jump Point Search to accelerate planning by pruning neighbors, reducing computational overhead. Tangent graph approaches have also evolved: Liu et al. (Liu and Arimoto 1992) constructed tangent graphs including all locally shortest paths for polygonal and curved obstacles, proving that lines on locally shortest paths must be tangent to obstacles. Yao et al. (Yao et al. 2019) suggested precomputing tangent graphs to reduce planning time, though computation remains costly for complex polygons. Cheriet et al. (Cheriet et al. 2025) proposed the Tangent Intersection Guidance (TIG) algorithm, a two-dimensional geometric approach based on elliptical obstacle modeling that improves planning speed and path smoothness compared to classical methods. The extended version, TIG* (CHERIET et al. 2026), generalizes the framework to three-dimensional environments by introducing prism-based obstacle modeling and refined sub-path selection rules, resulting in improved path quality while maintaining computational efficiency.

New intelligent optimization approaches enhance 3D planning by improving convergence and efficiency. These include PSO with adaptive weights and genetic operators (Liang et al. 2023), Levy-flight PSO (LGPSO) for urban navigation (Cheng et al. 2024), hybrid PSO-APF to avoid local minima (Mishra and Sevil 2024), and improved continuous ant colony optimization path planning in complex 3D environments (Niu et al. 2025). Despite their advantages, these methods remain sensitive to initialization, can easily get trapped in local minima, require careful parameter tuning, and become computationally expensive, especially for paths with many waypoints that force the planner to operate in high-dimensional spaces. They can also struggle with narrow passages or highly cluttered environments, limiting their reliability in real-world scenarios.

Machine learning techniques have recently been applied to overcome the limitations of classical and optimization-based planners. Neural A* (Yonetani et al. 2020) predicts cost-to-go or promising path regions using a trained network, guiding search toward near-optimal solutions and reducing unnecessary exploration. To reduce the number of collision checks, Diao et al. (Diao et al. 2024) proposed a learning-based path planning method that integrates a graph neural network, where edge weights are predicted to guide exploration and avoid unnecessary collision evaluations. However, relying solely on such predictions can be risky in complex or unseen environments, potentially leading to unsafe decisions.

Neural RRT* (Wang et al. 2020) and Neural Informed RRT* (Huang et al. 2024) use neural networks to identify high-value sampling regions, reducing convergence time and accelerating tree expansion. Neural RRT* generates a nonuniform sampling distribution from CNN predictions trained on A* paths, while Neural Informed RRT* combines learned guidance with informed ellipsoidal sampling and connectivity constraints. Both methods mix learned and uniform samples to preserve probabilistic completeness, but their performance may degrade in highly irregular environments due to training bias toward structured obstacle configurations and the reliance on extensive state-space sampling.

Complementary approaches such as Motion Planning Networks (MPNet) (Qureshi et al. 2021) and risk-aware planners such as NR-RRT (Meng et al. 2024b, 2024a) further explore direct trajectory prediction and uncertainty-aware decision-making. MPNet learns to recursively generate near-optimal trajectories by utilizing experience from expert demonstrations, enabling efficient planning in both seen and unseen environments. NR-RRT extends this idea by introducing a neural risk-aware sampling framework that explicitly considers collision probability in uncertain nonconvex environments, improving safety while maintaining near-optimal performance. However, these methods typically operate at the trajectory or state level, which can limit flexibility in geometrically complex environments.

While learning-based planners improve efficiency and success rates, they are limited by prediction errors, generalization to unseen environments, and the need for large training datasets. Hybridization with classical strategies or fallback mechanisms is often required to ensure safety and completeness, particularly in high-dimensional 3D spaces or cluttered environments with narrow passages.

To address these challenges, our approach combines neural-guided sampling with classical planning strategies to balance computational efficiency and path optimality. The neural model predicts high-value regions near optimal waypoints, but relying solely on this prediction can be costly and prone to errors in irregular environments. By integrating grid-based planning with convex-guided sampling, the planner exploits both global feasible regions and local promising areas, ensuring efficient tree expansion and robustness even when model predictions are imperfect.

# 3 Methodology

## 3.1 Problem Formulation

The path planning problem is defined in a two-dimensional workspace $X \subset \mathbb{R}^2$, which is divided into two regions: free space $X_{free}$ and occupied space $X_{obs}$. The free region represents collision-free areas where the robot can move, while $X_{obs}$ represents occupied regions. To ensure safety during navigation, all obstacles are inflated by a safety margin $r_c$ as follows:

$$X_{obs}^c = \{x \in X \mid d(x, X_{obs}) \leq r_c\}, \quad X_{free}^c = X \backslash X_{obs}^c,$$

where $d(x, X_{obs})$ denotes the minimum Euclidean distance between a point $x$ and the nearest obstacle cell.

A feasible path is a sequence of waypoints $P = \{p_1, p_2, \dots, p_N\}$, with $p_1 = x_{start}$ and $p_N = x_{goal}$, such that $p_i \in X_{free}^c$ for all $i$. The optimal path minimizes the total distance:

$$c(P) = \sum_{i=1}^{N-1} \| p_{i+1} - p_i \|, \qquad P^* = \underset{P \in \mathcal{P}}{\operatorname{argmin}}\, c(P),$$

where $\mathcal{P}$ is the set of all feasible, collision-free paths between $x_{start}$ and $x_{goal}$.

To evaluate the geometric quality of a path, we compute its smoothness using the cumulative change in heading between consecutive segments:

$$S = \sum_{i=1}^{N-1} |\theta_i - \theta_{i-1}|, \qquad \theta_i = \mathrm{atan2}(p_{i+1} - p_i).$$

Lower values of $S$ indicate smoother paths with fewer sharp turns, which are desirable for motion execution and control.

Beyond path quality, computational efficiency is a key consideration, particularly for sampling-based planners operating in cluttered environments. The geometric structure of obstacles plays an important role in determining both path optimality and exploration efficiency.

Obstacles in realistic environments can exhibit complex shapes. For analytical clarity, we distinguish between convex and concave obstacles. A convex obstacle $O \subset X_{obs}$ is defined such that, for any two points $x_1, x_2 \in O$, the line segment connecting them lies entirely within the obstacle:

$$\lambda x_1 + (1-\lambda)x_2 \in O, \quad \forall \lambda \in [0,1].$$

Convex obstacles simplify collision detection and related computations. On the other hand, a concave obstacle is defined as an obstacle $O \subset X_{obs}$ for which there exists at least one pair of points $x_1, x_2 \in O$ such that the line segment connecting them is not fully contained within $O$:

$$\exists x_1, x_2 \in O,\ \exists \lambda \in (0,1) \text{ s.t. } \lambda x_1 + (1-\lambda)x_2 \notin O.$$

As illustrated in Fig. 1, this distinction highlights structural differences in obstacle geometry.

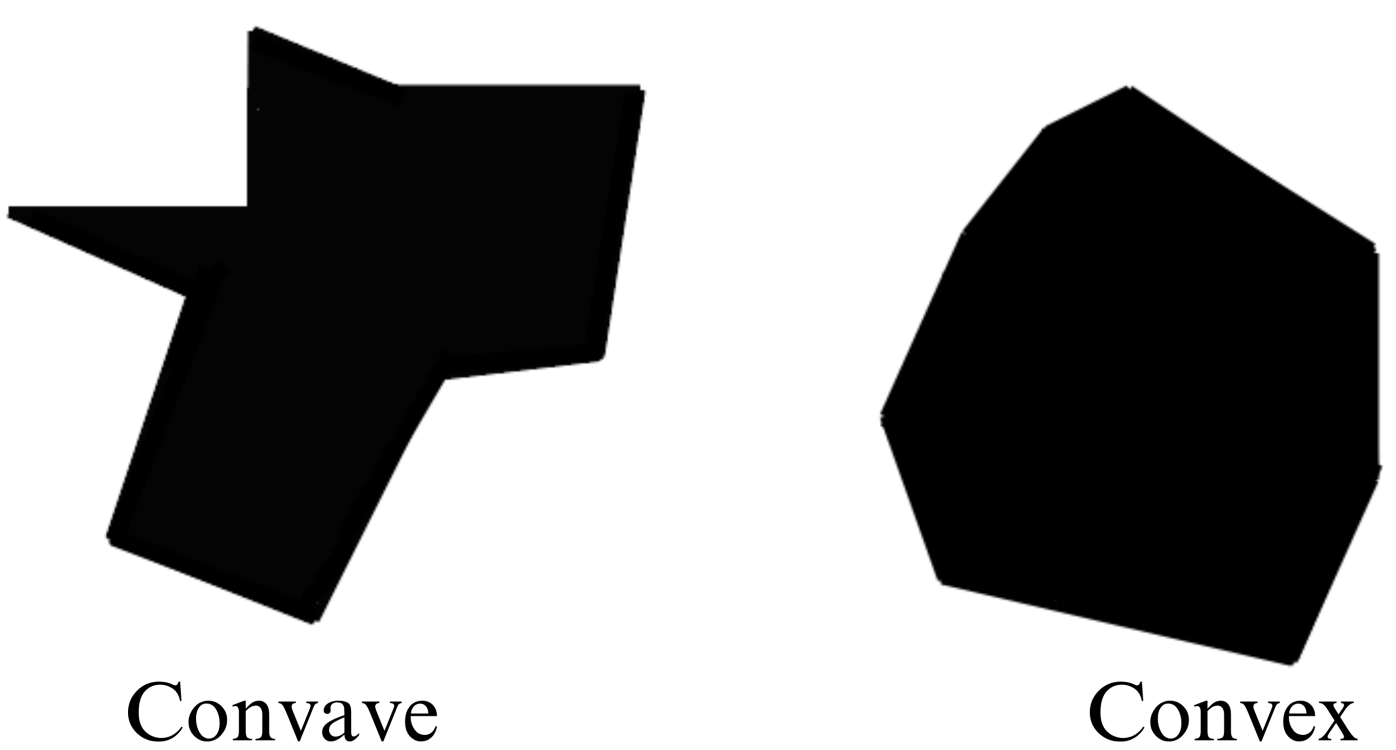

*Examples of concave and convex obstacles.*

Path planning algorithms that rely on uniform sampling over the free space do not explicitly incorporate this geometric structure. Because all free-space regions are treated with equal probability, structurally informative areas—such as regions near convex obstacle boundaries or visibility transitions—may not receive sufficient sampling density. This can lead to inefficient exploration, particularly in cluttered environments and narrow passages.

In this work, instead of relying solely on uniform random sampling over $X_{free}^{c}$, we introduce a structured sampling strategy derived from geometric obstacle features. Specifically, the sampling domain is restricted and guided by convex obstacle characteristics and learned waypoint predictions, as detailed in the following section.

## 3.2 Background on RRT*

The Rapidly-Exploring Random Tree (RRT) algorithm is a popular sampling-based planner for motion planning problems. The algorithm incrementally builds a tree of feasible configurations $\mathcal{T} = (\mathcal{V}, \mathcal{E})$, starting from the initial state $x_{\text{start}}$. At each iteration, a random sample $x_{\text{rand}} \in X_{\text{free}}$ is drawn, and the nearest node $x_{\text{nearest}} \in \mathcal{V}$ is selected according to a distance metric. A new node $x_{\text{new}}$ is generated by extending from $x_{\text{nearest}}$ toward $x_{\text{rand}}$ with a fixed step size $\delta$:

$$x_{\text{new}} = x_{\text{nearest}} + \delta \frac{x_{\text{rand}} - x_{\text{nearest}}}{\| x_{\text{rand}} - x_{\text{nearest}} \|}.$$

If the edge $(x_{\text{nearest}}, x_{\text{new}})$ is collision-free, $x_{\text{new}}$ is added to the tree. The process repeats until the goal $x_{\text{goal}}$ is reached or a predefined maximum iteration limit is exceeded.

While RRT efficiently explores high-dimensional spaces, it often produces suboptimal and jagged paths. To address this limitation, RRT* extends RRT by incorporating a rewiring mechanism that incrementally improves path quality. After adding a new node $x_{\text{new}}$, nearby nodes within a neighborhood radius $r_n$ are considered for optimal parent selection:

$$x_{\text{parent}}^{*} = \arg \min_{x \in \mathcal{X}_{\text{near}}} \{\text{cost}(x) + \| x - x_{\text{new}} \|\},$$

where $\text{cost}(x)$ denotes the cumulative path cost from $x_{\text{start}}$ to $x$. Subsequently, a rewiring step reconnects neighboring nodes if their path cost can be reduced through $x_{\text{new}}$. Through repeated sampling and rewiring, RRT* exhibits incremental improvement in solution quality.

The performance of RRT* strongly depends on its sampling strategy. In its standard formulation, samples are drawn uniformly from the free space without explicitly considering the geometric structure of the environment. Consequently, many samples may be allocated to regions that contribute little to improving the current best path. In structured environments, optimal or near-optimal paths are typically shaped by obstacle boundary characteristics, especially convex vertices that act as key visibility points. This insight motivates the introduction of a structured sampling strategy that prioritizes geometrically significant regions while maintaining the incremental refinement behavior of RRT*, as detailed in the next section.

## 3.3 Convex Corner Extraction

Obstacle boundaries naturally contain geometrically significant *decision points*, corresponding to locations where a feasible path may need to change direction to maintain obstacle clearance. In particular, convex boundary vertices often define visibility transitions and influence shortest-path structure. Extracting such convex corner points provides a compact and geometrically meaningful representation of the environment for structured sampling.

To identify convex corners in a binary occupancy grid, Zafar et al. (Zafar et al. 2021) propose a simple $3 \times 3$ local neighborhood rule. Let $C(i,j)$ denote the grid cell at position $(i,j)$. A cell is labeled as a convex corner if:

$$C(i,j) \in X^c_{free} \quad \text{and} \quad \sum_{(u,v)\in\mathcal{N}_8(i,j)} \mathbb{I}\left[C(u,v) \in X^c_{obs}\right] = 1,$$

where $\mathcal{N}_8(i,j)$ represents the 8-connected neighbors of $(i,j)$, and $\mathbb{I}[\cdot]$ is the indicator function. This condition identifies free cells that are adjacent to exactly one inflated obstacle cell, corresponding to outward-facing convex boundary configurations in the grid representation. Consistent with prior grid-based approaches, this definition approximates convex corners using local occupancy patterns rather than strict geometric criteria, enabling the extraction of boundary transition points relevant for guiding the path planning process.

The resulting convex corner set is denoted as:

$$\mathcal{C} = \{c_1, c_2, \dots, c_M \mid c_i \text{ satisfies the above condition}\}.$$

These extracted corners form a sparse set of structurally informative points that guide the sampling process in subsequent stages. An example of convex and concave corner identification on an obstacle boundary is shown in Fig. 2.

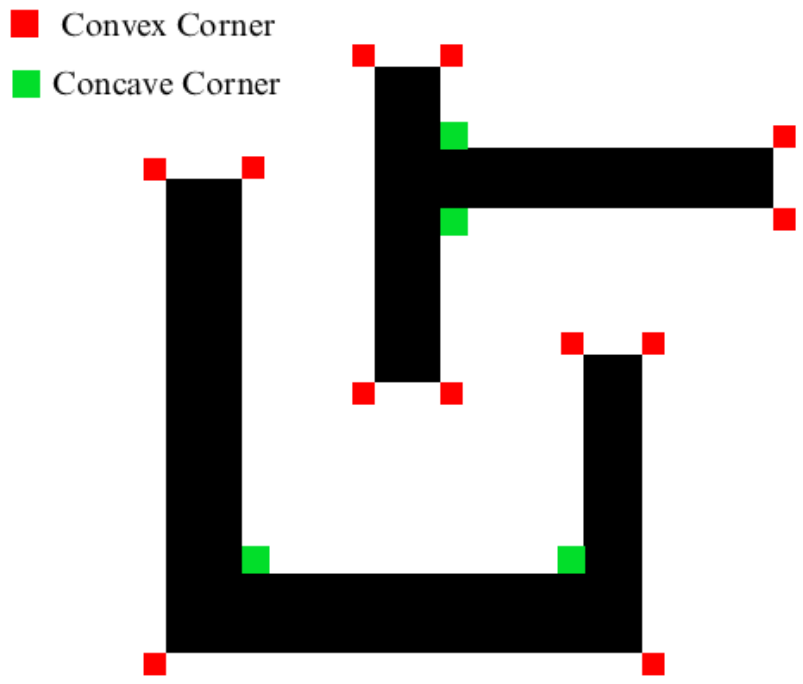

*Illustration of convex corner detection. Detected corners are highlighted in red along the obstacle boundaries.*

## 3.4 Neural-Guided Extensions of RRT*

To improve the exploration efficiency of RRT*, several learning-based variants have been proposed, including Neural RRT* and Neural Informed RRT*. These approaches incorporate neural network predictions into the sampling process in order to bias exploration toward promising regions of the environment, potentially reducing the number of iterations required to reach the goal and improving path quality. Neural RRT* combines neural-guided sampling with uniform random sampling. At each iteration, a sample $x_{\text{rand}}$ is selected according to a mixed strategy:

$$x_{\text{rand}} = \begin{cases} x_{\text{pred}}, & \text{with probability } \alpha, \\ x_{\text{uniform}}, & \text{with probability } 1-\alpha, \end{cases}$$

where $x_{\text{pred}}$ is sampled from the set of points predicted by the neural network as likely to lie along the optimal path, and $x_{\text{uniform}}$ is sampled uniformly from the free space $X_{\text{free}}$. The parameter $\alpha \in [0,1]$ controls the trade-off between guided sampling and random exploration. This mixed strategy preserves exploration capability even when predictions are imperfect.

Neural Informed RRT* further integrates learning with the admissible ellipsoidal sampling region of Informed RRT*. A Neural Connect module is introduced to promote connectivity between model-predicted regions and encourage consistent tree expansion. This integration enables more focused exploration while retaining the theoretical properties of RRT* under standard assumptions.

Despite these advancements, certain limitations remain. Neural RRT* still relies on continuous uniform sampling over the configuration space, which may allocate samples to regions with limited relevance to improving the current solution. In complex environments, the predicted guidance regions may be sparse or disconnected, potentially slowing convergence. Neural Informed RRT* addresses many of these issues, yet it still presents several limitations. Its performance depends heavily on the similarity between test environments and the training distribution, making generalization to substantially larger or more complex environments uncertain. The guidance points inferred by the point-based network may include noise, suggesting the need for further refinement. Moreover, Neural Connect introduces additional computational overhead, and the approach inherits the admissible ellipsoidal heuristic from Informed RRT*, which can be less effective under more complex structures. These limitations highlight the importance of integrating learning-based guidance with explicit geometric reasoning, which forms the foundation of the convex-guided sampling strategy introduced in the next section.

## 3.5 Convex-Neural RRT*

To address the limitations of existing neural-guided sampling methods, we propose **Convex-Neural RRT***, a structured sampling framework that integrates convex obstacle geometry with neural waypoint prediction within the RRT* planning process. The proposed approach combines the convex-hull guidance principle inspired by LTA* with learning-based region prediction to bias exploration toward geometrically informative areas while preserving the incremental refinement mechanism of RRT*.

## 3.6 Model Settings

Unlike previous approaches that rely on global heatmap prediction, the proposed method learns *promising regions* rather than reconstructing an entire feasible path, which can lead to fragmented or disconnected guidance. In our framework, these regions are derived from a visibility-graph construction that connects convex obstacle corners within the grid map, providing a geometrically interpretable and computationally efficient structure for learning-based guidance.

Accordingly, the network is trained to predict inflated region masks generated by expanding intermediate waypoints obtained from LTA* paths. These regions form a structured supervision signal, while convex-corner points are extracted from the predicted regions during post-processing.

Similar to Neural RRT*, a dataset is constructed from multiple grid environments with varying start and goal positions. Each training sample consists of a grid map represented using three input channels encoding free space, occupied space, and start/goal locations, paired with a corresponding binary region mask generated from LTA* paths by extracting intermediate waypoints and constructing inflated region masks around them, excluding the start and goal cells. To improve training stability and reduce label sparsity, each waypoint in the ground-truth mask is inflated by a radius of six grid cells. This inflation enlarges the supervision signal while preserving the underlying geometric structure.

The resulting input and output tensors are of size $224 \times 224 \times 3$ and $224 \times 224 \times 1$, respectively (see Fig. 3). The predictor employs a U-Net architecture with a ResNet50 encoder pretrained on ImageNet. The encoder remains frozen and is used as a fixed feature extractor, while the decoder is trained to generate the output masks. The network therefore learns to identify convex-corner-based promising regions that subsequently guide the RRT* sampling process toward geometrically relevant areas of the map.

For reproducibility, we summarize the dataset and training configuration used for this auxiliary neural module. A total of 4,000 grid maps were procedurally generated, each containing a mixture of convex and concave polygonal obstacles. For each map, ten distinct feasible start–goal pairs were assigned, resulting in 40 000 labeled samples. Because the ground-truth masks are highly sparse (less than 1% positive pixels), class-balancing weights were applied during training. The model was trained for 100 epochs using the Adam optimizer with a learning rate of $10^{-3}$, a batch size of 128, and cross-entropy loss.

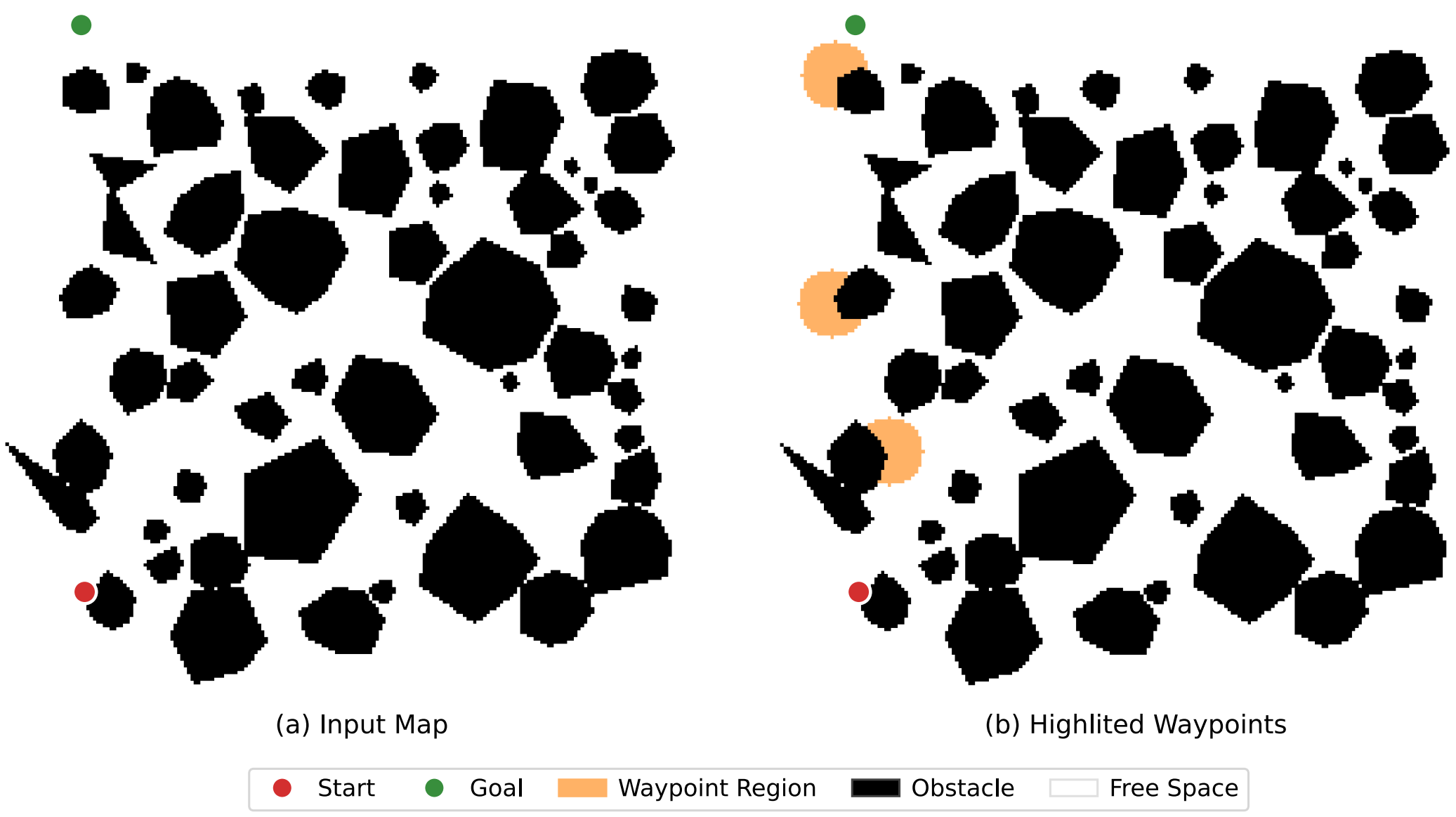

*Input grid map and its corresponding region mask generated from LTA* waypoint expansion*

### 3.6.1 Design of the Convex-Neural RRT*

To improve computational efficiency while preserving the incremental refinement behavior of RRT*, let $\mathcal{C}_p$ denote the convex-corner waypoints extracted from the predicted mask, expanded by a small radius $r_c$ to compensate for localization uncertainty. A convex hull $H_c$ is constructed over the union of these predicted convex corners, the start, and the goal positions. This hull defines a geometrically informed subregion that reduces the effective search space while maintaining connectivity between structurally relevant regions. As illustrated in Fig. 4, the model predicts promising regions (yellow). After applying a filtering step, only convex-corner points that lie within these predicted regions are retained (highlighted in red), forming the set $\mathcal{C}_p$.

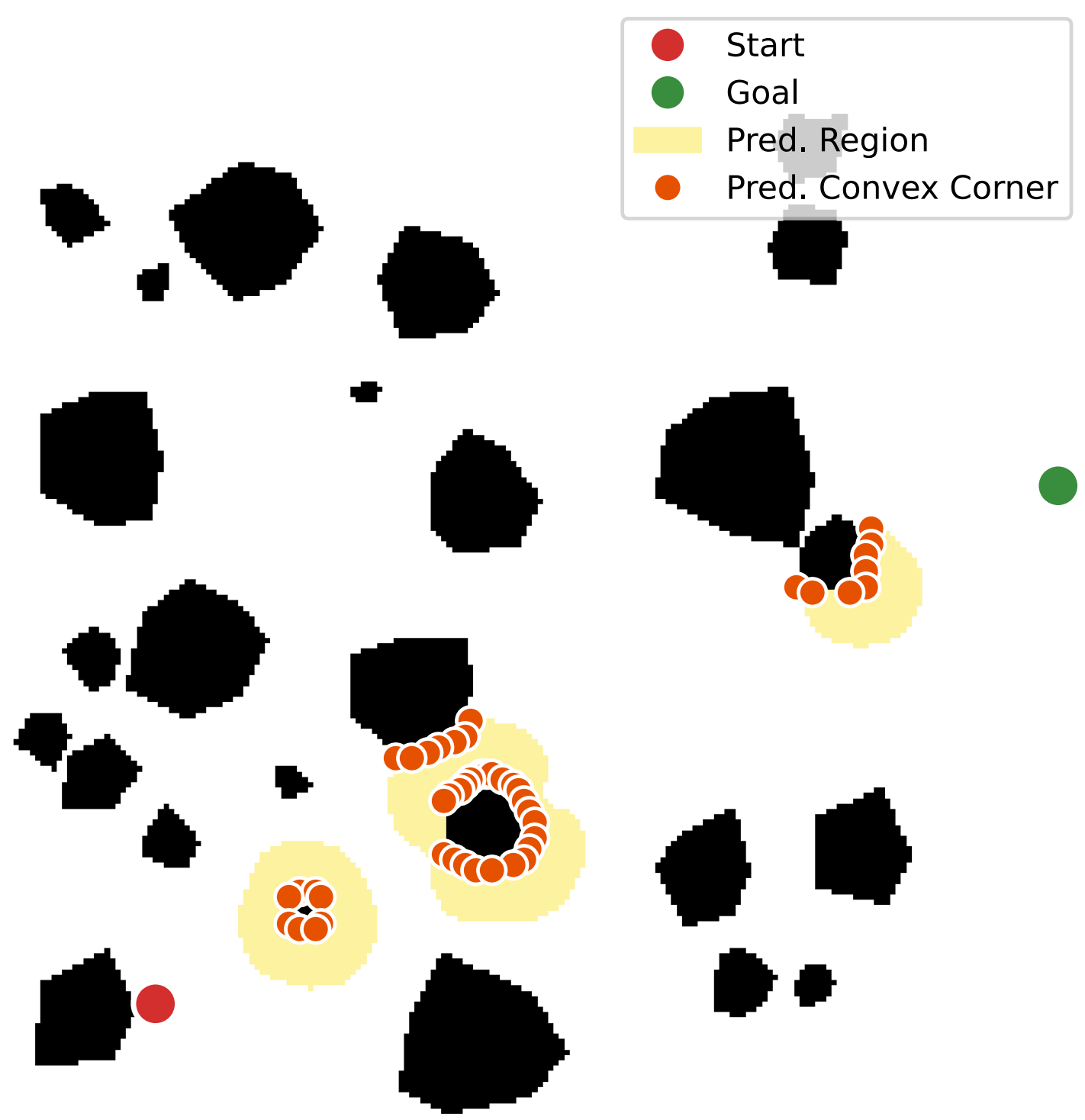

*Predicted promising regions (yellow) are filtered to keep only the convex-corner waypoints contained within them (red).*

Instead of sampling uniformly over the entire free space, Convex-Neural RRT* performs structured discrete sampling over convex-corner sets. Samples are drawn from three subsets: predicted convex corners inside the hull $\mathcal{C}_p$, remaining convex corners within the hull $\mathcal{C}_{ri}$, and convex corners outside the hull $\mathcal{C}_{ro}$. To regulate the balance between exploitation and exploration, two independent parameters are introduced: $\alpha_{\text{pred}}$ and $\alpha_{\text{explore}}$. - $\alpha_{\text{pred}}$ governs the probability of sampling from predicted convex-corner points relative to other convex corners inside $H_c$. - $\alpha_{\text{explore}}$ controls the probability of sampling outside the hull, enabling global exploration.

The overall random sampling strategy is represented as:

$$\begin{equation} \footnotesize \mathbf{x}_{rand} = \begin{cases} \text{random}(\mathcal{C}_{ro}), & \text{with probability } \alpha_{\text{explore}},\\[1mm] \text{random}(\mathcal{C}_{p}), & \text{with probability } (1-\alpha_{\text{explore}})\,\alpha_{\text{pred}},\\[1mm] \text{random}(\mathcal{C}_{ri}), & \text{with probability } (1-\alpha_{\text{explore}})\,(1-\alpha_{\text{pred}}), \end{cases} \end{equation}$$

This probabilistic formulation enables focused exploitation of predicted promising regions while preserving sufficient exploration capacity to mitigate prediction errors and maintain path feasibility. Fig. 5 illustrates the resulting tree expansion. Most nodes concentrate within the convex hull due to the influence of $\alpha_{\text{pred}}$, while a smaller fraction extends outside the hull to sustain global coverage. Tree expansion follows the standard RRT* procedure. For each sampled target, the nearest node is identified, and a new node is generated by steering toward the target with step size $\delta$, subject to collision checking to ensure that connections between nodes are feasible and safe. A local rewiring step then evaluates nearby nodes and reassigns parents when cost reduction is possible, thereby progressively improving path quality. Because sampling is restricted to a predefined convex-corner set, an early-stopping criterion is introduced to avoid unnecessary expansion once convergence stabilizes. The search terminates when the best path cost does not improve over the last $t_s$ iterations. Let $c_k$ denote the best cost at iteration $k$; early stopping is triggered when

$$|c_k - c_{k-t_s}| < \varepsilon,$$

indicating that additional iterations are unlikely to yield significant improvement.

The proposed method retains the core structure of RRT* while introducing a structured sampling strategy. The algorithm combines guided sampling over convex-corner sets with controlled exploration to maintain coverage of the search space and support consistent convergence behavior.

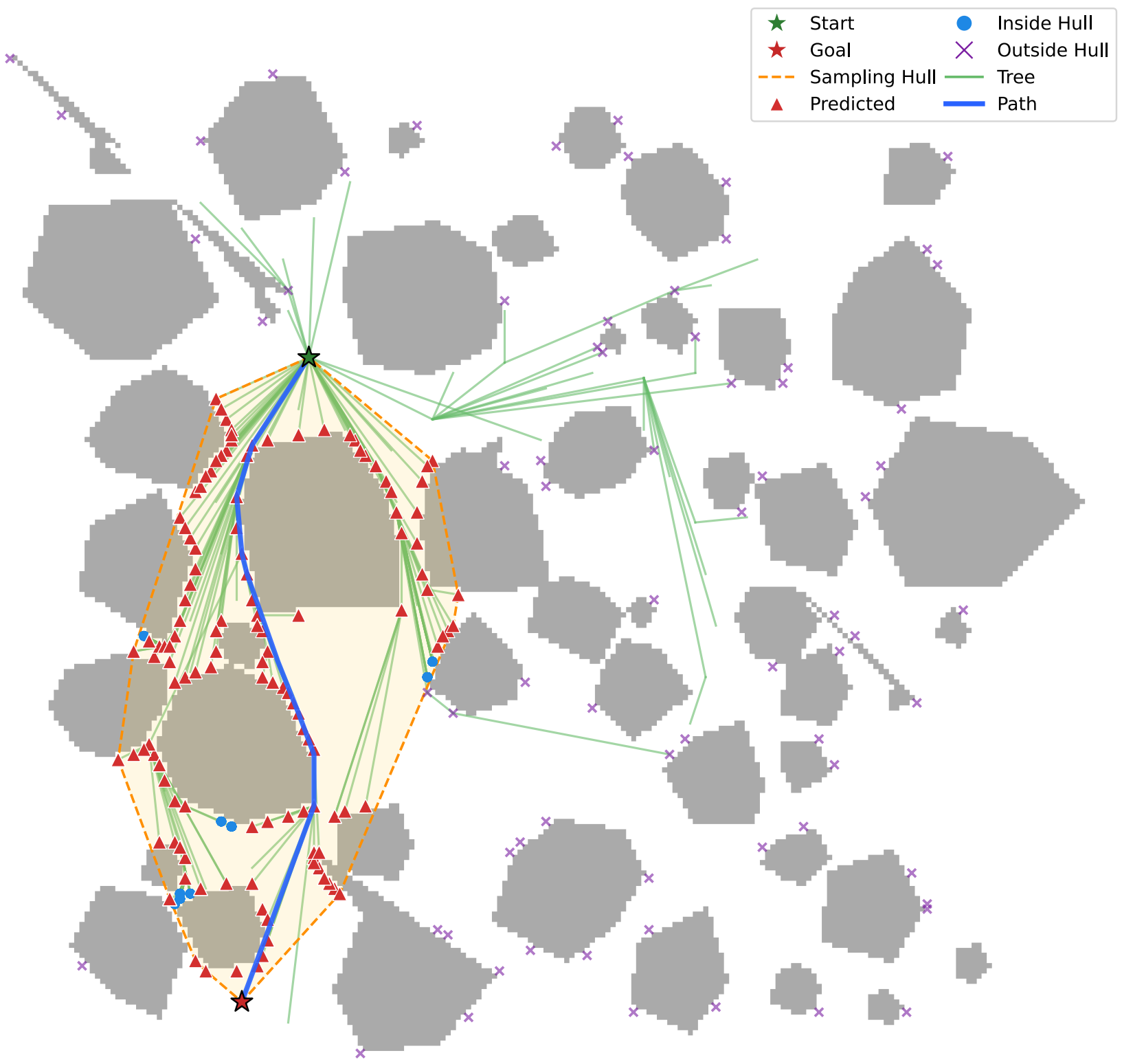

*Convex-Neural RRT*: predicted convex-corner points (red) define a sampling hull (orange) that guides focused sampling and tree expansion toward the goal.*

The complete procedure of the Convex-Neural RRT* is summarized in Algorithm [alg:convex_rrt_star].

Grid $G$, start $S$, goal $T$, predicted convex-corner points $\mathcal{C}_p$, real convex-corner points $\mathcal{C}_r$ $\alpha_{\text{pred}}$, $\alpha_{\text{explore}}$, $\delta_s$, $N_{max}$, $t_s$

Initialize node list $\mathcal{V} \leftarrow \{S\}$, cost($S$) $\leftarrow 0$

Compute convex hull $H_c \leftarrow$ ConvHull($\mathcal{C}_p \cup \{S, T\}$)

$\mathcal{C}_{ri} \leftarrow \mathcal{C}_r \backslash \mathcal{C}_p$ $\mathcal{C}_{ro} \leftarrow$ points in $\mathcal{C}_r$ located outside $H_c$

Sample $r \sim U[0,1]$

$$\boldsymbol{x}_{rand} \leftarrow \text{random}(\mathcal{C}_{ro})$$

$$\boldsymbol{x}_{rand} \leftarrow \text{random}(\mathcal{C}_{p})$$

$$\boldsymbol{x}_{rand} \leftarrow \text{random}(\mathcal{C}_{ri})$$

$\boldsymbol{x}_{near} \leftarrow$ Nearest($\mathcal{V}, \boldsymbol{x}_{rand}$) $\boldsymbol{x}_{new} \leftarrow$ Steer($\boldsymbol{x}_{near}, \boldsymbol{x}_{rand}, \delta_s$)

Connect $\boldsymbol{x}_{new}$ to cost-optimal parent in neighborhood $\mathcal{N}$ Rewire nodes in $\mathcal{N}$ if cost decreases Update best goal cost $c_k$

**break**

Extract path from $T$ to $S$

## 3.7 Complexity Analysis

Standard RRT* has an expected computational complexity of $O(n\log n)$ due to nearest-neighbor queries and local rewiring operations, where $n$ denotes the number of nodes in the tree. In practice, achieving high-quality solutions may require a large number of iterations, particularly in cluttered or high-dimensional environments.

In Convex-Neural RRT*, sampling is restricted to a predefined set of convex-corner points derived from geometric structure and neural predictions. Let $\mathcal{C}_p$ denote predicted convex corners and $\mathcal{C}_r$ denote real convex corners. Because sampling is performed over this structured and typically sparse set rather than the entire free space, the effective exploration region is reduced. Specifically:

- Sampling is confined to convex-corner subsets inside and outside the convex hull, limiting unnecessary exploration of geometrically irrelevant regions,
- The convex-hull restriction concentrates tree growth within an informative subregion, reducing the dispersion of nodes,
- An early-stopping criterion based on cost stabilization limits the effective number of iterations once convergence behavior is observed.

Although the theoretical worst-case complexity remains comparable to RRT*, the structured sampling strategy reduces the practical number of required iterations. Let $N_{\text{eff}}$ denote the effective number of iterations before early stopping, as determined by the convergence criterion in Eq. ([eq:early_stop]), which terminates the search once the path cost stabilizes; in practice, $N_{\text{eff}} \ll n$ for many environments. Therefore, the empirical computational cost can be approximated as

$$O(N_{\text{eff}} \log N_{\text{eff}}),$$

reflecting the reduced exploration horizon.

This structured reduction in effective sampling and iteration count leads to improved practical efficiency, as observed in the experimental results. The analysis is therefore empirical in nature and reflects the observed behavior of the proposed sampling strategy.

# 4 Results and Discussion

To evaluate the efficiency and effectiveness of the proposed Convex-Neural RRT* algorithm, we compare it with four baseline methods: classical RRT*, Neural RRT*, Neural Informed RRT*, and LTA*. The evaluation considers both computational efficiency and final path quality under identical experimental conditions. In particular, each algorithm is assessed in terms of final path length, smoothness, execution time, and success rate. These metrics collectively capture path optimality, maneuverability, computational cost, and reliability, providing a comprehensive comparison across multiple experimental environments. The experimental setup is described below.

## 4.1 Environment Settings

Three types of experimental environments were designed on a $224 \times 224$ grid map, each representing a different level of planning difficulty:

- **Sparse-density environment:** represents an open navigation scenario with widely spaced obstacles. This case evaluates the algorithm's ability to exploit large free-space regions and generate efficient paths in minimally cluttered maps.
- **Medium-density environment:** represents a general navigation scenario where obstacles are randomly generated with low to moderate density. This case evaluates algorithm performance under typical operating conditions.
- **High-density environment:** contains a larger number of obstacles with reduced free space. This scenario enables a more challenging comparison between planners and evaluates their ability to find feasible solutions in cluttered maps.

For fair comparison, all algorithms were tested using identical start and goal positions for each map. Obstacles were modeled as randomly generated convex and concave polygons. The step size was set to $\eta = 5$, the neighbor radius to $r_{near} = 7$, and the maximum number of iterations to $N_{iter} = 1000$. The sampling ratio ($\alpha$) for Neural RRT* and Neural Informed RRT* was set to $0.5$. For Convex-Neural RRT*, $\alpha_{\text{explore}} = 0.2$ and $\alpha_{\text{pred}} = 0.5$.

All learning-based planners (Neural RRT*, Neural Informed RRT*, and Convex-Neural RRT*) use the same U-Net–based architecture and training configuration. The difference lies only in the supervision labels: Neural RRT* and Neural Informed RRT* predict dense full-path regions, whereas Convex-Neural RRT* predicts sparse waypoint regions (inflated convex-corner blobs). This difference is inherent to the underlying planning strategies, as dense path prediction and region-based guidance represent fundamentally different ways of guiding sampling rather than differences in model capacity or training conditions, ensuring a fair comparison under matched architectural settings.

All experiments were implemented in Python using a Kaggle notebook and executed on the default Kaggle CPU configuration (Intel Xeon CPU at 2.2 GHz, 30 GB RAM, no GPU acceleration). These hardware and software conditions were kept consistent across all experiments.

Neural predictions were obtained from two pre-trained models evaluated at the same grid resolution. The first model predicts dense path regions and guides Neural RRT* and Neural Informed RRT*. The second model predicts convex-corner waypoint regions for Convex-Neural RRT*. Both models were trained on the same dataset to ensure comparable prediction quality. The distinction in output representation—dense path regions versus sparse convex waypoint regions—enables Convex-Neural RRT* to focus exploration on geometrically informative areas while maintaining global exploration capability.

## 4.2 Comparison Results

To compare efficiency and effectiveness, each algorithm was executed 10 times per map, and performance metrics were collected, including path length, execution time, smoothness, and success rate. For all metrics, the mean and standard deviation are reported, reflecting both typical performance and variability across repeated runs. The results are summarized in Tabs. [tab:results_easy], [tab:results_medium], and [tab:results_hard]. Figs. 6–8 illustrate representative paths consistent with these results.

| Map | Algorithm | Length ($m$) | Time ($s$) | Smoothness | Success (%) |
|---|---|---|---|---|---|
| 1 | RRT* | $214.16 \pm 5.87$ | $0.54 \pm 0.03$ | $5.10 \pm 1.47$ | 100.0 |
| | Neural RRT* | $209.85 \pm 1.88$ | $0.50 \pm 0.08$ | $4.21 \pm 1.08$ | 100.0 |
| | Neural Inf RRT* | $207.73 \pm 2.27$ | $0.42 \pm 0.03$ | $3.34 \pm 0.98$ | 100.0 |
| | LTA* | $\mathbf{203.19 \pm 0.00}$ | $1.54 \pm 0.03$ | $1.64 \pm 0.00$ | 100.0 |
| | Convex-Neural RRT* | $205.57 \pm 5.90$ | $\mathbf{0.11 \pm 0.05}$ | $\mathbf{1.42 \pm 0.46}$ | 100.0 |
| 2 | RRT* | $135.68 \pm 0.65$ | $0.58 \pm 0.09$ | $2.11 \pm 0.69$ | 100.0 |
| | Neural RRT* | $135.50 \pm 0.54$ | $0.52 \pm 0.06$ | $2.20 \pm 0.37$ | 100.0 |
| | Neural Inf RRT* | $134.26 \pm 0.33$ | $0.47 \pm 0.01$ | $1.39 \pm 0.54$ | 100.0 |

| Map | Algorithm | Length ($m$) | Time ($s$) | Smoothness | Success (%) |
|---|---|---|---|---|---|
| | LTA* | $\mathbf{133.46 \pm 0.00}$ | $0.96 \pm 0.02$ | $\mathbf{0.31 \pm 0.00}$ | 100.0 |
| | Convex-Neural RRT* | $133.61 \pm 0.20$ | $\mathbf{0.05 \pm 0.02}$ | $0.32 \pm 0.02$ | 100.0 |
| 3 | RRT* | $192.90 \pm 0.08$ | $0.53 \pm 0.05$ | $0.10 \pm 0.10$ | 100.0 |
| | Neural RRT* | $193.23 \pm 0.64$ | $0.53 \pm 0.06$ | $0.24 \pm 0.24$ | 100.0 |
| | Neural Inf RRT* | $192.86 \pm 0.01$ | $0.46 \pm 0.02$ | $0.04 \pm 0.05$ | 100.0 |
| | LTA* | $192.86 \pm 0.00$ | $0.17 \pm 0.00$ | $\mathbf{0.01 \pm 0.00}$ | 100.0 |
| | Convex-Neural RRT* | $\mathbf{192.86 \pm 0.01}$ | $\mathbf{0.04 \pm 0.01}$ | $\mathbf{0.01 \pm 0.02}$ | 100.0 |
| 4 | RRT* | $174.05 \pm 0.04$ | $0.56 \pm 0.10$ | $0.05 \pm 0.07$ | 100.0 |
| | Neural RRT* | $174.06 \pm 0.08$ | $0.62 \pm 0.10$ | $0.07 \pm 0.14$ | 100.0 |
| | Neural Inf RRT* | $174.03 \pm 0.00$ | $0.47 \pm 0.02$ | $0.02 \pm 0.00$ | 100.0 |
| | LTA* | $174.03 \pm 0.00$ | $0.14 \pm 0.00$ | $\mathbf{0.00 \pm 0.00}$ | 100.0 |
| | Convex-Neural RRT* | $\mathbf{174.03 \pm 0.00}$ | $\mathbf{0.00 \pm 0.00}$ | $\mathbf{0.00 \pm 0.00}$ | 100.0 |
| 5 | RRT* | $182.45 \pm 1.49$ | $0.53 \pm 0.05$ | $4.16 \pm 1.01$ | 100.0 |
| | Neural RRT* | $182.58 \pm 2.04$ | $0.54 \pm 0.06$ | $4.01 \pm 1.13$ | 100.0 |
| | Neural Inf RRT* | $179.57 \pm 0.74$ | $0.45 \pm 0.03$ | $2.71 \pm 0.66$ | 100.0 |
| | LTA* | $177.38 \pm 0.00$ | $0.77 \pm 0.02$ | $0.48 \pm 0.00$ | 100.0 |
| | Convex-Neural RRT* | $\mathbf{177.35 \pm 0.03}$ | $\mathbf{0.10 \pm 0.05}$ | $\mathbf{0.42 \pm 0.13}$ | 100.0 |
| 6 | RRT* | $157.12 \pm 0.39$ | $0.54 \pm 0.03$ | $0.50 \pm 0.29$ | 100.0 |
| | Neural RRT* | $157.01 \pm 0.27$ | $0.52 \pm 0.03$ | $0.66 \pm 0.48$ | 100.0 |
| | Neural Inf RRT* | $156.70 \pm 0.03$ | $0.50 \pm 0.03$ | $0.25 \pm 0.09$ | 100.0 |
| | LTA* | $156.67 \pm 0.00$ | $0.26 \pm 0.01$ | $0.09 \pm 0.00$ | 100.0 |
| | Convex-Neural RRT* | $\mathbf{156.64 \pm 0.00}$ | $\mathbf{0.05 \pm 0.02}$ | $\mathbf{0.08 \pm 0.01}$ | 100.0 |

| Map | Algorithm | Length ($m$) | Time ($s$) | Smoothness | Success (%) |
|---|---|---|---|---|---|
| 1 | RRT* | $200.95 \pm 9.29$ | $0.39 \pm 0.05$ | $8.88 \pm 2.83$ | 100.0 |
| | Neural RRT* | $208.14 \pm 12.00$ | $0.44 \pm 0.06$ | $10.05 \pm 2.28$ | 100.0 |
| | Neural Inf RRT* | $193.70 \pm 6.02$ | $0.38 \pm 0.04$ | $7.81 \pm 1.55$ | 100.0 |
| | LTA* | $183.63 \pm 0.00$ | $7.78 \pm 0.13$ | $3.24 \pm 0.00$ | 100.0 |
| | Convex-Neural RRT* | $\mathbf{182.77 \pm 1.62}$ | $\mathbf{0.23 \pm 0.09}$ | $\mathbf{2.59 \pm 0.82}$ | 100.0 |
| 2 | RRT* | $177.71 \pm 0.86$ | $0.44 \pm 0.05$ | $2.20 \pm 0.69$ | 100.0 |
| | Neural RRT* | $177.49 \pm 0.87$ | $0.48 \pm 0.08$ | $2.87 \pm 0.81$ | 100.0 |
| | Neural Inf RRT* | $176.14 \pm 0.41$ | $0.47 \pm 0.01$ | $1.61 \pm 0.42$ | 100.0 |
| | LTA* | $\mathbf{175.00 \pm 0.00}$ | $1.31 \pm 0.02$ | $0.26 \pm 0.00$ | 100.0 |
| | Convex-Neural RRT* | $175.01 \pm 0.05$ | $\mathbf{0.13 \pm 0.04}$ | $\mathbf{0.25 \pm 0.01}$ | 100.0 |
| 3 | RRT* | $115.85 \pm 0.26$ | $0.46 \pm 0.05$ | $1.13 \pm 0.48$ | 100.0 |

| Map | Algorithm | Length ($m$) | Time ($s$) | Smoothness | Success (%) |
|---|---|---|---|---|---|
| | Neural RRT* | $115.75 \pm 0.20$ | $0.51 \pm 0.07$ | $0.97 \pm 0.26$ | 100.0 |
| | Neural Inf RRT* | $115.32 \pm 0.06$ | $0.50 \pm 0.01$ | $0.56 \pm 0.10$ | 100.0 |
| | LTA* | $\mathbf{115.28 \pm 0.00}$ | $0.63 \pm 0.01$ | $\mathbf{0.15 \pm 0.00}$ | 100.0 |
| | Convex-Neural RRT* | $115.35 \pm 0.17$ | $\mathbf{0.04 \pm 0.02}$ | $0.17 \pm 0.03$ | 100.0 |
| 4 | RRT* | $177.50 \pm 22.72$ | $0.49 \pm 0.07$ | $6.88 \pm 3.83$ | 100.0 |
| | Neural RRT* | $164.84 \pm 4.39$ | $0.52 \pm 0.06$ | $5.00 \pm 1.68$ | 100.0 |
| | Neural Inf RRT* | $161.34 \pm 0.97$ | $0.45 \pm 0.05$ | $3.24 \pm 0.72$ | 100.0 |
| | LTA* | $\mathbf{159.52 \pm 0.00}$ | $3.86 \pm 0.02$ | $0.71 \pm 0.00$ | 100.0 |
| | Convex-Neural RRT* | $159.69 \pm 0.32$ | $\mathbf{0.18 \pm 0.06}$ | $\mathbf{0.66 \pm 0.16}$ | 100.0 |
| 5 | RRT* | $132.90 \pm 0.97$ | $0.53 \pm 0.10$ | $2.47 \pm 0.51$ | 100.0 |
| | Neural RRT* | $132.93 \pm 0.71$ | $0.48 \pm 0.06$ | $2.28 \pm 0.94$ | 100.0 |
| | Neural Inf RRT* | $131.81 \pm 0.58$ | $0.47 \pm 0.03$ | $1.81 \pm 0.43$ | 100.0 |
| | LTA* | $\mathbf{131.10 \pm 0.00}$ | $3.83 \pm 0.04$ | $\mathbf{0.75 \pm 0.00}$ | 100.0 |
| | Convex-Neural RRT* | $131.45 \pm 1.15$ | $\mathbf{0.10 \pm 0.04}$ | $0.77 \pm 0.11$ | 100.0 |
| 6 | RRT* | $159.52 \pm 0.81$ | $0.52 \pm 0.06$ | $1.85 \pm 0.53$ | 100.0 |
| | Neural RRT* | $159.34 \pm 1.04$ | $0.51 \pm 0.06$ | $2.03 \pm 0.58$ | 100.0 |
| | Neural Inf RRT* | $158.07 \pm 0.33$ | $0.47 \pm 0.02$ | $1.46 \pm 0.35$ | 100.0 |
| | LTA* | $\mathbf{157.63 \pm 0.00}$ | $1.77 \pm 0.03$ | $\mathbf{0.60 \pm 0.00}$ | 100.0 |
| | Convex-Neural RRT* | $157.76 \pm 0.32$ | $\mathbf{0.06 \pm 0.03}$ | $0.60 \pm 0.11$ | 100.0 |

| Map | Algorithm | Length ($m$) | Time ($s$) | Smoothness | Success (%) |
|---|---|---|---|---|---|
| 1 | RRT* | $128.46 \pm 2.74$ | $0.31 \pm 0.09$ | $7.48 \pm 1.15$ | 90.0 |
| | Neural RRT* | $133.17 \pm 13.09$ | $0.30 \pm 0.07$ | $7.81 \pm 1.53$ | 100.0 |
| | Neural Inf RRT* | $124.62 \pm 4.53$ | $0.34 \pm 0.06$ | $4.85 \pm 1.17$ | 100.0 |
| | LTA* | $\mathbf{121.89 \pm 0.00}$ | $7.96 \pm 0.07$ | $\mathbf{3.31 \pm 0.00}$ | 100.0 |
| | Convex-Neural RRT* | $129.60 \pm 4.53$ | $\mathbf{0.29 \pm 0.15}$ | $4.30 \pm 1.17$ | 100.0 |
| 2 | RRT* | $235.97 \pm 38.13$ | $\mathbf{0.20 \pm 0.02}$ | $16.65 \pm 3.60$ | 60.0 |
| | Neural RRT* | $220.48 \pm 27.34$ | $0.21 \pm 0.04$ | $12.88 \pm 4.40$ | 60.0 |
| | Neural Inf RRT* | $198.51 \pm 11.58$ | $0.23 \pm 0.04$ | $10.71 \pm 2.54$ | 100.0 |
| | LTA* | $\mathbf{178.92 \pm 0.00}$ | $10.44 \pm 0.05$ | $3.30 \pm 0.00$ | 100.0 |
| | Convex-Neural RRT* | $182.12 \pm 4.89$ | $0.42 \pm 0.15$ | $\mathbf{3.28 \pm 1.37}$ | 100.0 |
| 3 | RRT* | $182.20 \pm 36.88$ | $0.23 \pm 0.05$ | $9.51 \pm 3.59$ | 100.0 |
| | Neural RRT* | $184.65 \pm 23.09$ | $\mathbf{0.21 \pm 0.06}$ | $8.96 \pm 2.79$ | 100.0 |
| | Neural Inf RRT* | $174.38 \pm 23.76$ | $0.29 \pm 0.08$ | $8.04 \pm 2.10$ | 100.0 |
| | LTA* | $\mathbf{149.84 \pm 0.00}$ | $3.89 \pm 0.03$ | $\mathbf{3.11 \pm 0.00}$ | 100.0 |

| Map | Algorithm | Length ($m$) | Time ($s$) | Smoothness | Success (%) |
|---|---|---|---|---|---|
| | Convex-Neural RRT* | $170.36 \pm 26.14$ | $0.47 \pm 0.14$ | $5.48 \pm 2.16$ | 100.0 |
| 4 | RRT* | $181.03 \pm 9.27$ | $\mathbf{0.25 \pm 0.04}$ | $11.40 \pm 2.60$ | 80.0 |
| | Neural RRT* | $173.28 \pm 5.04$ | $0.31 \pm 0.04$ | $10.32 \pm 1.36$ | 80.0 |
| | Neural Inf RRT* | $170.67 \pm 5.06$ | $0.32 \pm 0.03$ | $9.79 \pm 1.35$ | 90.0 |
| | LTA* | $165.18 \pm 0.00$ | $19.98 \pm 0.18$ | $\mathbf{4.04 \pm 0.00}$ | 100.0 |
| | Convex-Neural RRT* | $\mathbf{165.04 \pm 6.33}$ | $0.39 \pm 0.15$ | $6.32 \pm 1.36$ | 90.0 |
| 5 | RRT* | $132.46 \pm 4.34$ | $0.33 \pm 0.03$ | $8.09 \pm 1.41$ | 100.0 |
| | Neural RRT* | $136.35 \pm 8.16$ | $0.32 \pm 0.06$ | $8.97 \pm 1.33$ | 100.0 |
| | Neural Inf RRT* | $127.16 \pm 2.69$ | $0.33 \pm 0.04$ | $6.56 \pm 1.26$ | 100.0 |
| | LTA* | $\mathbf{124.24 \pm 0.00}$ | $12.87 \pm 0.06$ | $\mathbf{4.11 \pm 0.00}$ | 100.0 |
| | Convex-Neural RRT* | $128.34 \pm 6.05$ | $\mathbf{0.27 \pm 0.13}$ | $4.68 \pm 1.01$ | 100.0 |
| 6 | RRT* | $171.68 \pm 29.66$ | $0.23 \pm 0.09$ | $7.72 \pm 2.48$ | 100.0 |
| | Neural RRT* | $161.97 \pm 24.46$ | $0.22 \pm 0.07$ | $7.31 \pm 2.66$ | 100.0 |
| | Neural Inf RRT* | $146.39 \pm 1.58$ | $0.28 \pm 0.11$ | $4.88 \pm 0.77$ | 100.0 |
| | LTA* | $\mathbf{142.70 \pm 0.00}$ | $6.80 \pm 0.07$ | $\mathbf{3.70 \pm 0.00}$ | 100.0 |
| | Convex-Neural RRT* | $147.53 \pm 3.26$ | $\mathbf{0.19 \pm 0.12}$ | $4.24 \pm 0.66$ | 100.0 |

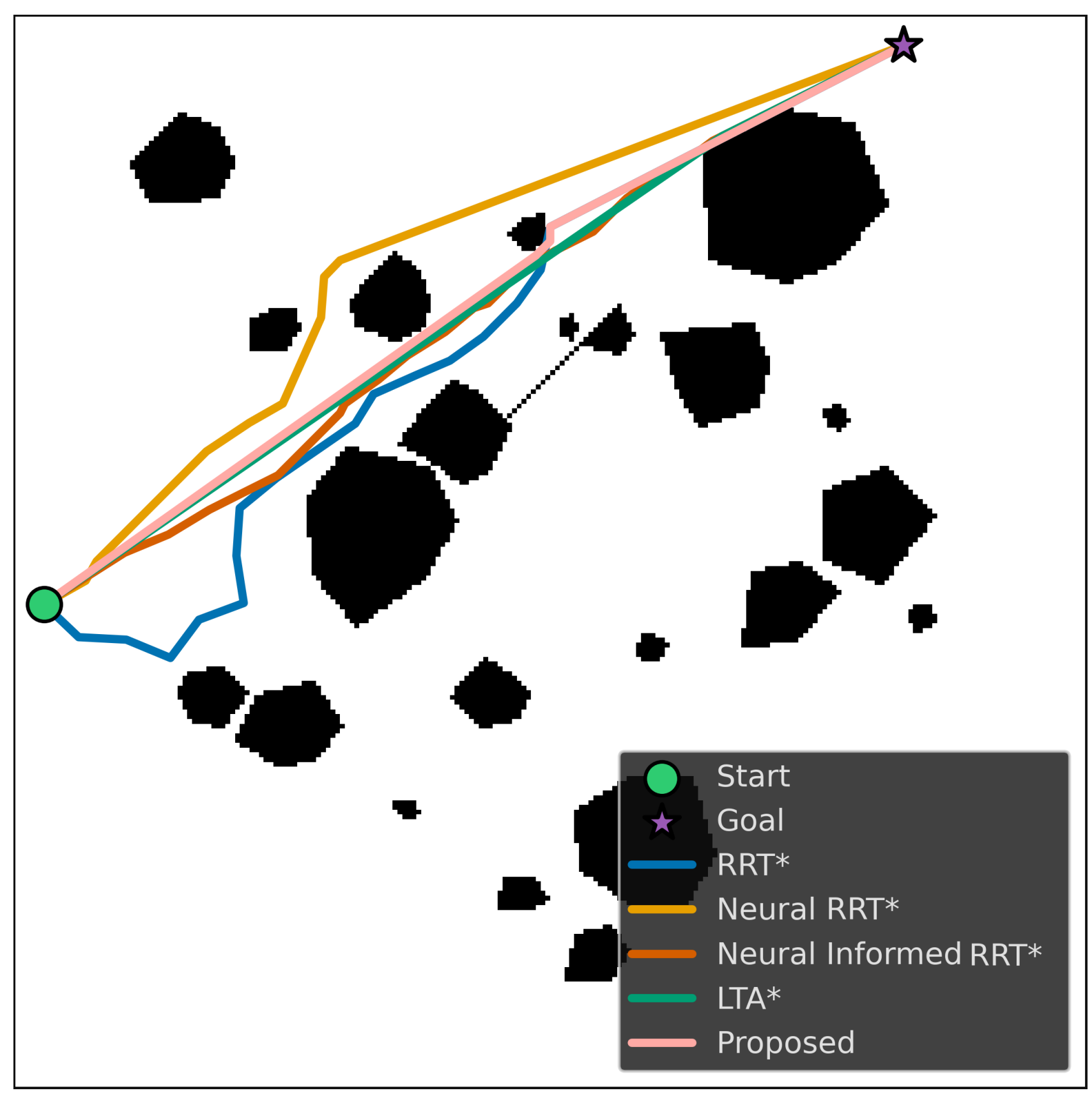

*Planned Paths Comparison (Sparse-Density Map)*

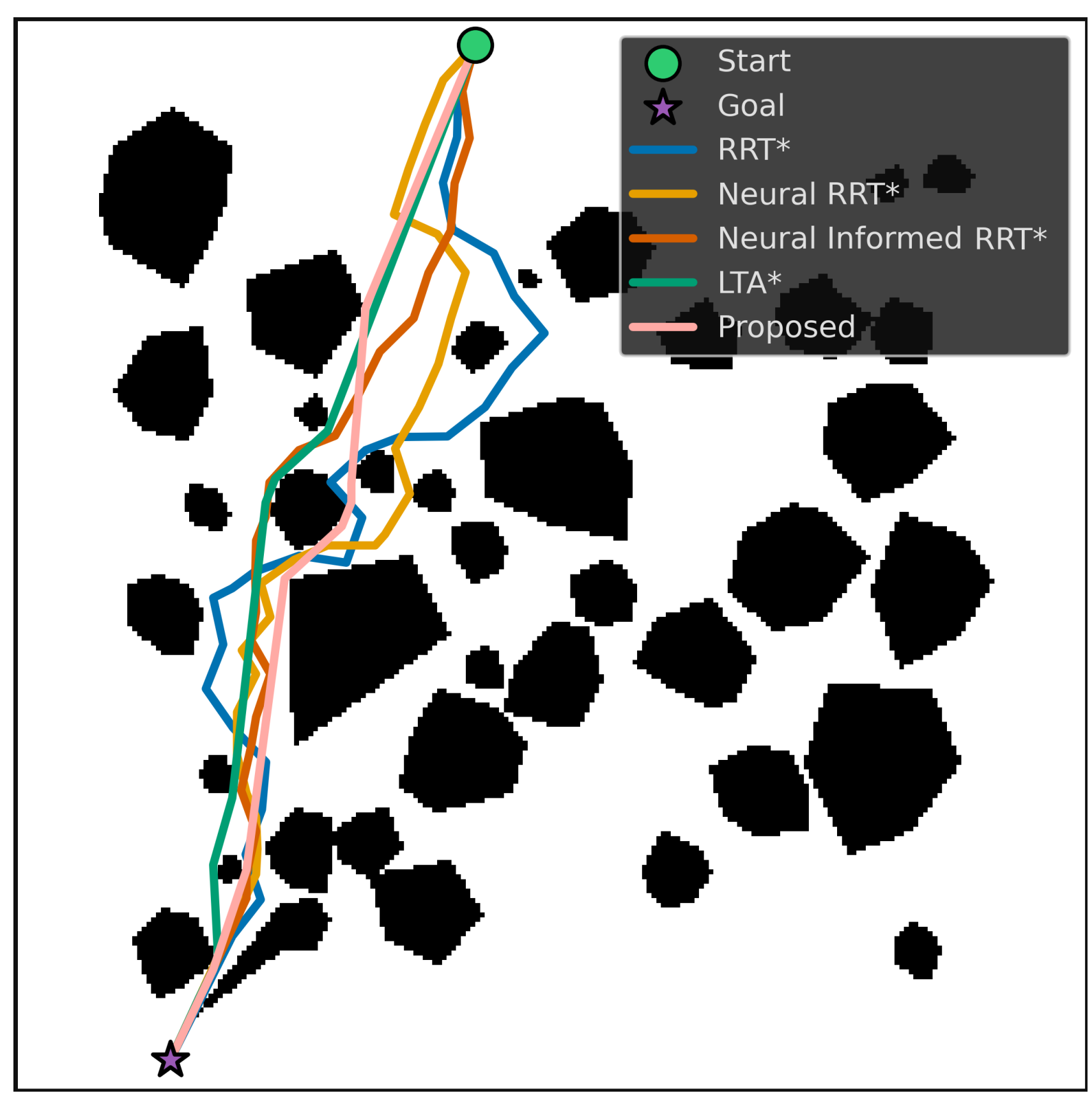

*Planned Paths Comparison (Medium-Density Map)*

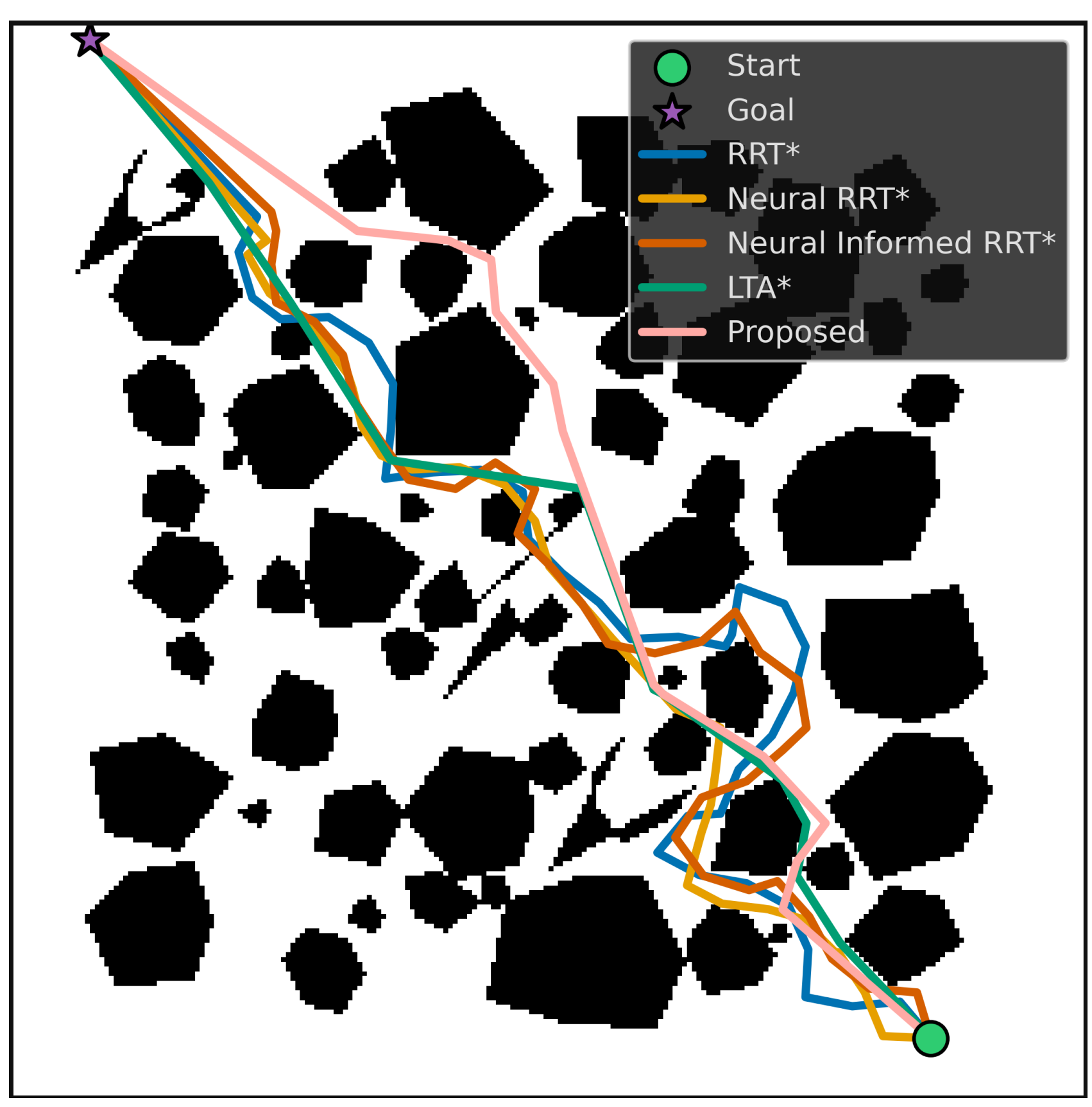

*Planned Paths Comparison (High-Density Map)*

### 4.2.1 Path Length

Across most maps, Convex-Neural RRT* produced shorter paths than classical RRT* and Neural RRT*, and achieved competitive performance relative to Neural Informed RRT*. On average, it reduced path length by 5.49% relative to classical RRT*, with improvements of 10.55% in Hard maps, 4.39% in Medium maps, and 1.55% in Easy maps. The most significant gain was observed on Hard Map 1, where RRT* generated a path of 235.97 m compared to 182.12 m for Convex-Neural RRT*, corresponding to a reduction of approximately 22.8%.

Neural RRT* and Neural Informed RRT* also generally outperformed classical RRT*, achieving average reductions of approximately 1.05% and 4.24%, respectively. Compared to Neural RRT*, Convex-Neural RRT* further shortened paths by an average of 4.49%, while the improvement over Neural Informed RRT* was more modest at 1.30%. These results suggest that convex-corner waypoint prediction provides additional geometric guidance beyond heuristic neural steering alone.

LTA* generated the shortest paths in most cases. Nevertheless, Convex-Neural RRT* remained close to its performance, with an average difference of 1.49%. The difference was negligible in Easy and Medium maps (0.24% and 0.01%, respectively), while the gap increased in Hard maps (4.55%).

Path length variance for Convex-Neural RRT* was generally low in Easy and Medium environments (standard deviations often below 2 m), but higher variability was observed in several Hard maps, particularly Hard Map 2 (26.14 m) and Hard Map 3 (6.33 m), reflecting the increased sensitivity of sampling-based planners in cluttered environments.

### 4.2.2 Execution Time

Regarding execution time, Convex-Neural RRT* substantially reduced computational cost compared to both LTA* and classical RRT*. Across the majority of tested maps, execution time for Convex-Neural RRT* remained below 0.5 s, whereas LTA* required up to 19.98 s on highly complex maps. This corresponds to a reduction of approximately 88–98% in Medium and Hard maps, and 70–100% in Easy maps.

Compared to Neural Informed RRT*, Convex-Neural RRT* was generally 30–75% faster in most Easy and Medium environments, although in several Hard maps the execution times were comparable or slightly higher.

Furthermore, the consistently low standard deviation of Convex-Neural RRT* (typically around or below 0.15 s in most scenarios) indicates stable performance across repeated trials, supporting its suitability for time-sensitive navigation tasks.

The observed reduction in execution time can be attributed to three main factors. First, sampling is restricted to a sparse set of convex-corner points, resulting in fewer ineffective samples compared to uniform exploration. Second, the structured sampling strategy focuses the search on geometrically informative regions, improving the efficiency of tree expansion. Third, an early-stopping mechanism terminates the search once cost stabilization is detected, avoiding unnecessary iterations after convergence.

### 4.2.3 Smoothness

Path smoothness is an important metric in robot path planning, as it reflects trajectories with fewer abrupt direction changes. Reducing sharp turns can decrease energy consumption and improve motion stability. Convex-Neural RRT* consistently achieved smoother paths than classical RRT*, with improvements becoming more pronounced in Medium and Hard environments. On average across all maps, Convex-Neural RRT* reduced the smoothness cost by approximately 56% relative to classical RRT*, with the largest improvements observed in Hard maps, where reductions frequently exceeded 40–60%.

Compared to Neural RRT*, Convex-Neural RRT* achieved an average smoothness improvement of approximately 58%, while the improvement over Neural Informed RRT* was approximately 42%. These results suggest that convex-corner waypoint prediction reduces unnecessary curvature and promotes smoother trajectories.

Although LTA* generally generated the smoothest paths, Convex-Neural RRT* remained close to its performance in Easy and Medium maps and moderately higher in Hard maps. For example, in Hard Map 3, Convex-Neural RRT* achieved a smoothness value of $6.32 \pm$ $1.36$ compared to $4.04 \pm 0.00$ for LTA* and $11.40 \pm 2.60$ for classical RRT*, indicating a substantial improvement over sampling-based baselines while maintaining competitive performance relative to LTA*.

### 4.2.4 Success Rate

LTA* achieved a 100% success rate across all trials. In contrast, classical RRT* exhibited reduced reliability in several Hard maps, with success rates dropping to 60–90%. Neural RRT* and Neural Informed RRT* also showed occasional failures in Hard environments, with success rates decreasing to 60–80% and around 90%, respectively, in the most challenging scenarios. Neural RRT* may fail when its learned sampling distribution misguides exploration, while Neural Informed RRT* can be affected when prediction inaccuracies interact with the restricted informed-sampling region.

Convex-Neural RRT* achieved an overall success rate of approximately 99.4% across all experiments, with only one map (Hard Map 3) exhibiting a reduced success rate of 90%. This indicates that while the method remains highly reliable, performance in highly cluttered environments can be sensitive to the balance between predicted convex-region sampling and exploration outside the hull. These results highlight the importance of appropriately tuning $\alpha_{\text{pred}}$ and $\alpha_{\text{explore}}$ to maintain robustness in complex scenarios.

These results indicate that Convex-Neural RRT* effectively integrates neural-guided exploration with convex-corner waypoint prediction to bias the search toward high-quality solutions. Compared to classical RRT* and neural-guided variants, it achieves substantial reductions in computational cost while maintaining near-optimal path quality. The method consistently produces paths close to those of LTA*, particularly in Easy and Medium environments, while achieving significantly faster execution times. This balance between efficiency, smoothness, and reliability suggests promising potential for extension to real-time UAV and 3D path planning applications in diverse environments.

## 4.3 Parameter Sensitivity Analysis

To evaluate the robustness of Convex-Neural RRT* with respect to its sampling parameters $\alpha_{\text{pred}}$ and $\alpha_{\text{explore}}$, a sensitivity analysis was conducted. For this study, 10 randomly generated maps were used per difficulty level (sparse, medium, and hard). Each parameter configuration was executed 20 times per map, and the mean performance was recorded in terms of path length, computation time, smoothness, and success rate.

In sparse environments (Fig. 9), the planner exhibited low sensitivity to parameter variation. Except for the extreme case $\alpha_{\text{pred}} = 0$ and $\alpha_{\text{explore}} = 0$, where the success rate dropped to approximately 50%, all other configurations achieved nearly 100% success. Path length and smoothness showed only minor variations, indicating that either neural guidance or mild exploration is sufficient in simple environments.

In medium-density environments (Fig. 10), parameter sensitivity became more pronounced. When $\alpha_{\text{pred}} = 0$, the success rate degraded significantly (as low as 10%), demonstrating the importance of neural-guided sampling in moderately cluttered environments. Increasing $\alpha_{\text{pred}}$ consistently improved the success rate and path quality. The best trade-off was observed near $\alpha_{\text{pred}} = 0.9$ with $\alpha_{\text{explore}} = 0.1$, achieving 100% success with improved path length and smoothness.

In hard environments (Fig. 11), the influence of $\alpha_{\text{pred}}$ became critical. Without neural guidance, the success rate dropped as low as 2%. As $\alpha_{\text{pred}}$ increased, success rate improved substantially, reaching 100% for $\alpha_{\text{pred}} = 0.9$ with small exploration values. A small exploration term ($\alpha_{\text{explore}} = 0.1$) improved robustness compared to $\alpha_{\text{explore}} = 0$, while larger values slightly degraded smoothness and path length.

The parameter values used in the main experiments ($\alpha_{\text{pred}} = 0.5$, $\alpha_{\text{explore}} = 0.2$) were selected as a balanced configuration that generalizes across different environment types, rather than being optimized for a specific scenario. Although higher values of $\alpha_{\text{pred}}$ can improve performance in highly cluttered environments, they may reduce exploration capability and lead to decreased robustness in simpler or unseen maps. The chosen parameters, therefore, provide a trade-off between exploitation of predicted regions and global exploration, ensuring consistent performance across varying levels of complexity. In practice, these parameters can be adjusted depending on environment characteristics, with higher $\alpha_{\text{pred}}$ and lower $\alpha_{\text{explore}}$ preferred in dense environments, and more balanced values suitable for general use.

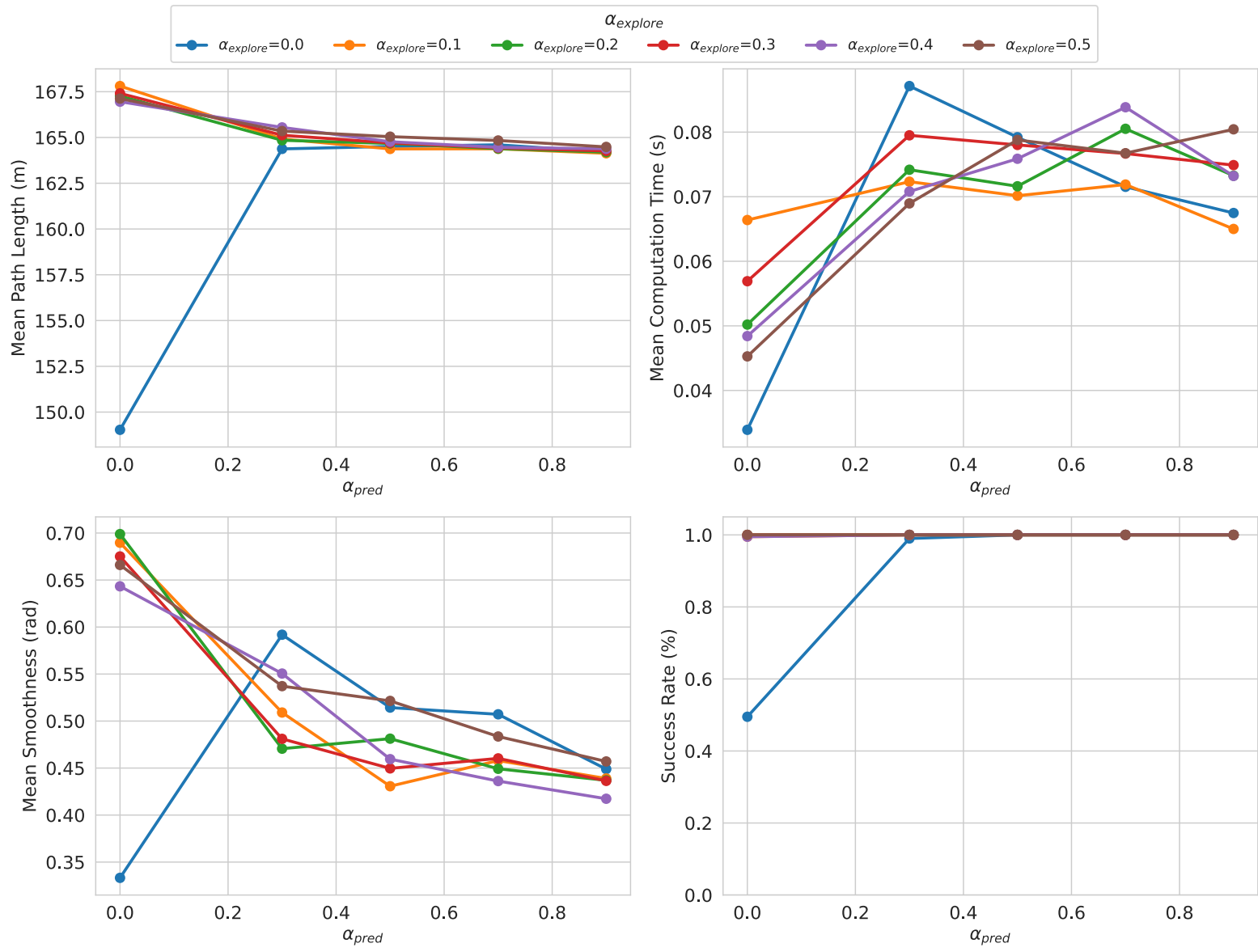

*Sensitivity analysis of Convex-Neural RRT* in sparse environments.*

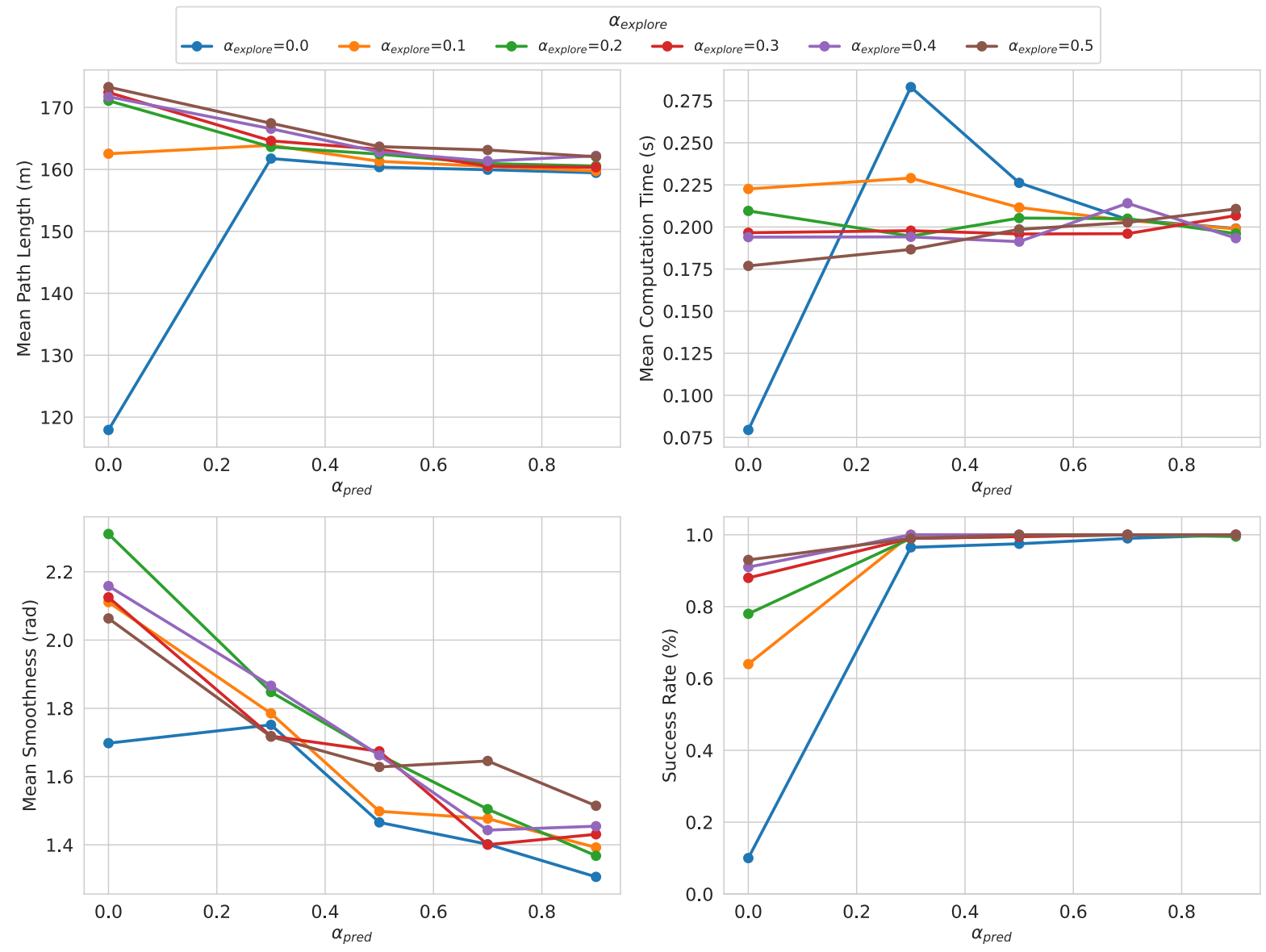

*Sensitivity analysis of Convex-Neural RRT* in medium environments.*

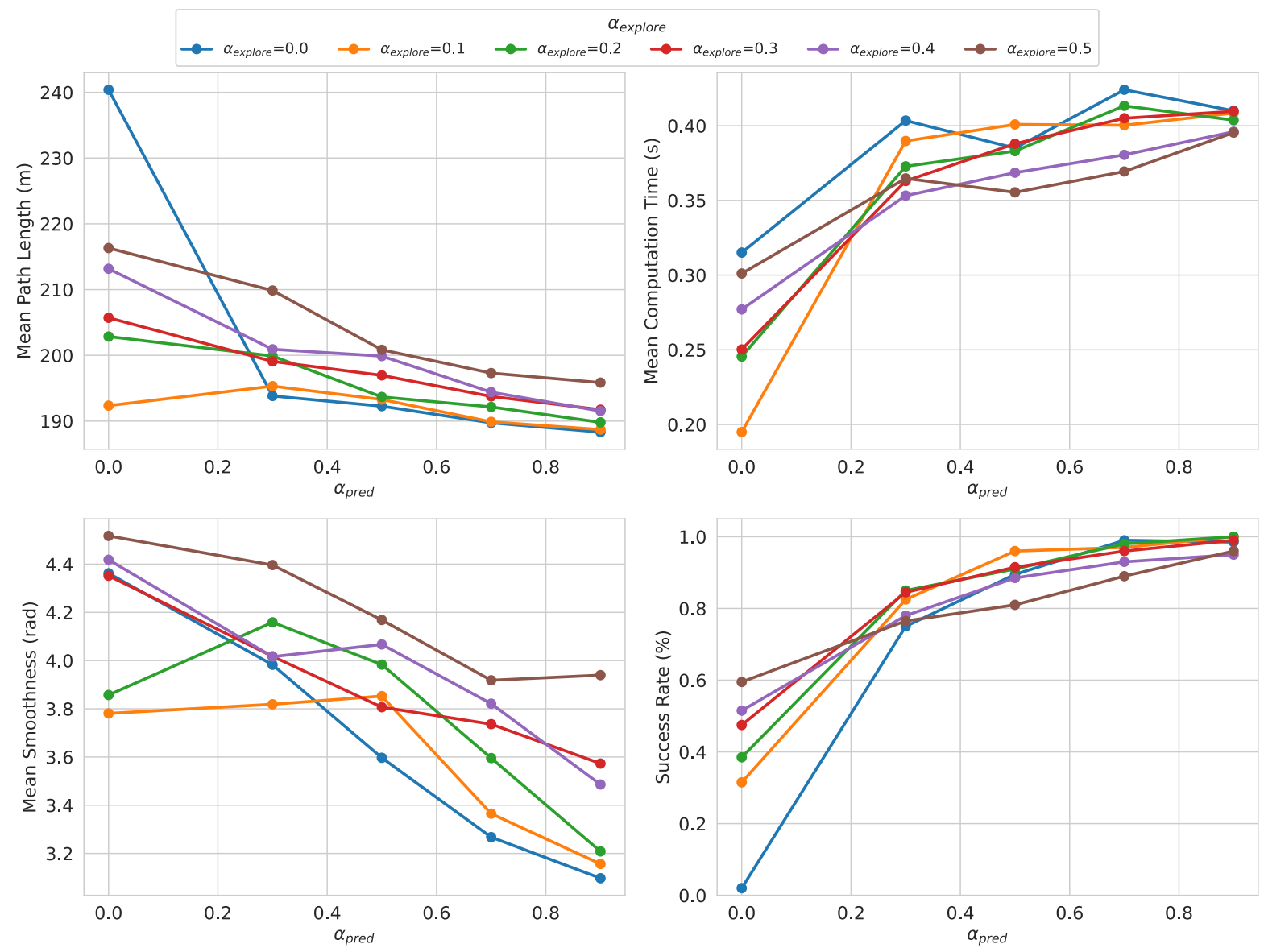

*Sensitivity analysis of Convex-Neural RRT* in dense environments.*

## 4.4 Convergence Analysis

To further evaluate the efficiency of the proposed algorithm, a convergence analysis was conducted in comparison with RRT*, Neural RRT*, and Neural Informed RRT*.

For each representative map (easy, medium, and dense), the best path cost obtained at each iteration was recorded and averaged over multiple runs.

The results illustrated in Figs. 12 to 14 show that the Convex-Neural RRT* planner consistently achieves faster cost reduction and earlier stabilization compared to the other approaches. In easy environments, all methods converge relatively quickly due to the lower obstacle density; however, Convex-Neural RRT* reaches a stable near-optimal solution using fewer iterations.

In medium and dense environments, the advantage becomes more pronounced. Classical RRT* exhibits slower improvement, while Neural RRT* and Neural Informed RRT* benefit from prediction guidance but still require additional iterations to refine the solution. In contrast, Convex-Neural RRT* demonstrates accelerated convergence by combining neural-guided sampling with convex-corner structural constraints.

This behavior supports the effectiveness of the proposed sampling strategy and further justifies the early-stopping mechanism, as the algorithm reaches cost stabilization significantly earlier than the maximum iteration limit.

Overall, the convergence analysis indicates that Convex-Neural RRT* not only improves final path quality but also reduces computational effort by attaining high-quality solutions more rapidly.

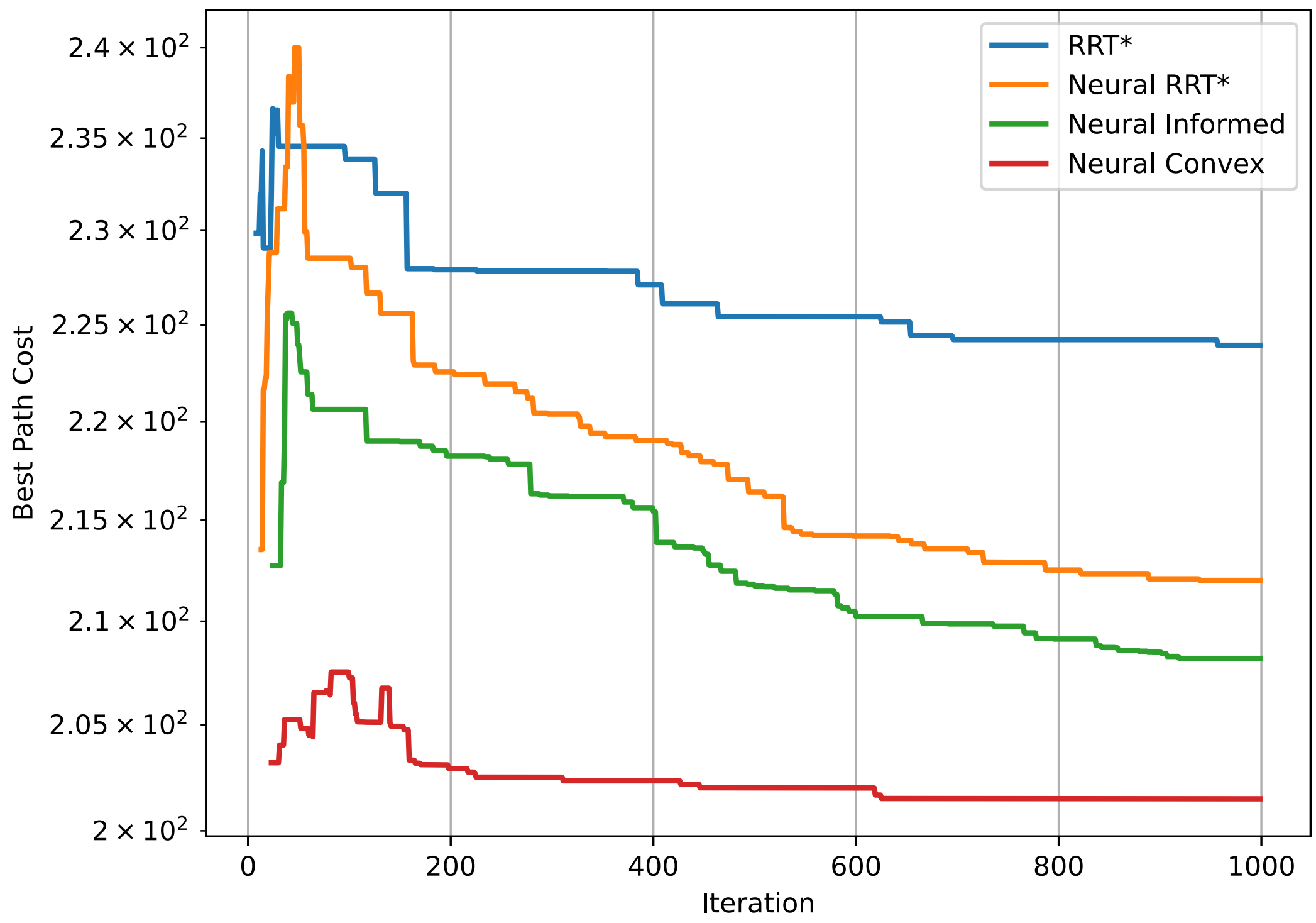

*Convergence curves on easy environments.*

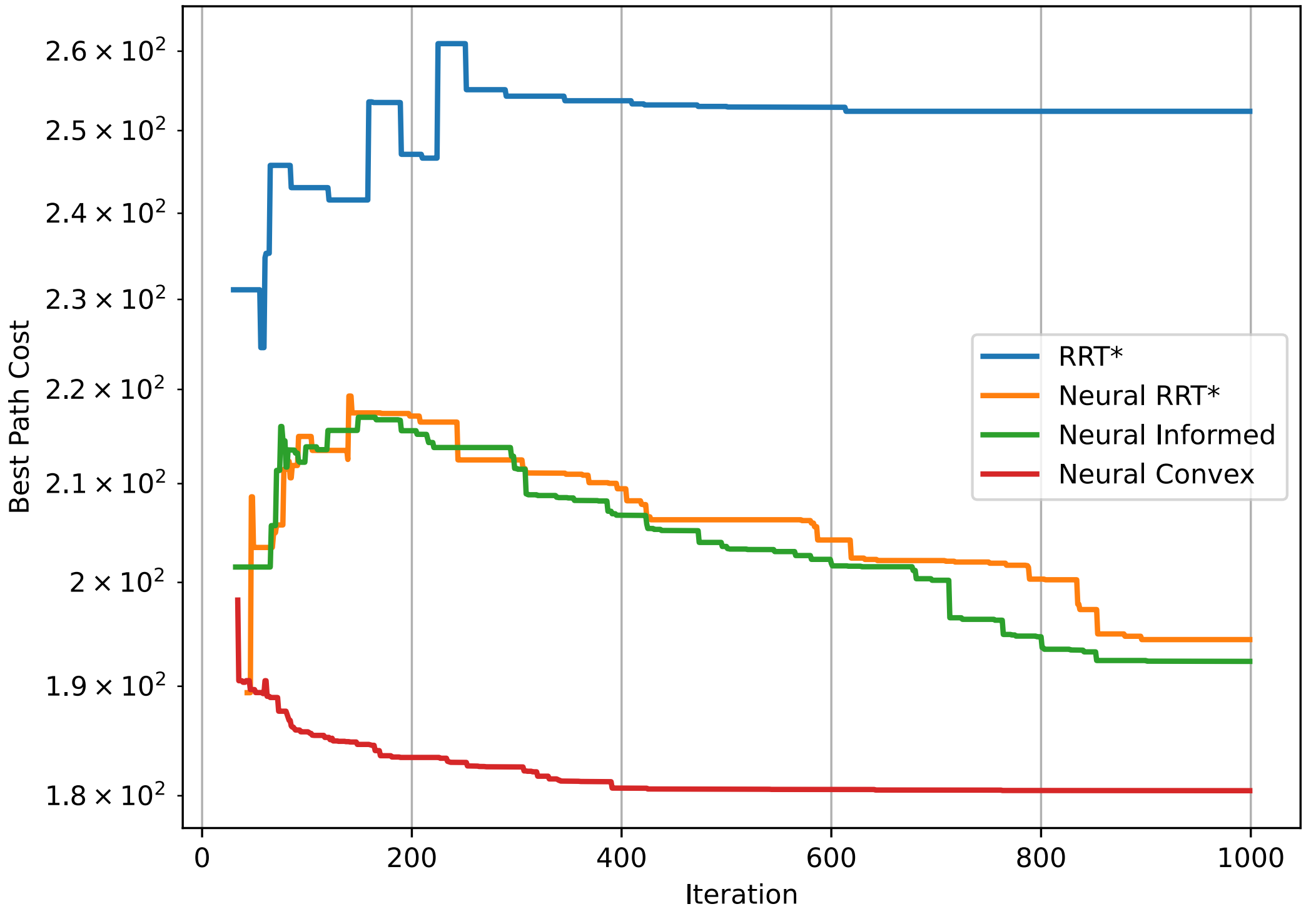

*Convergence curves on medium environments.*

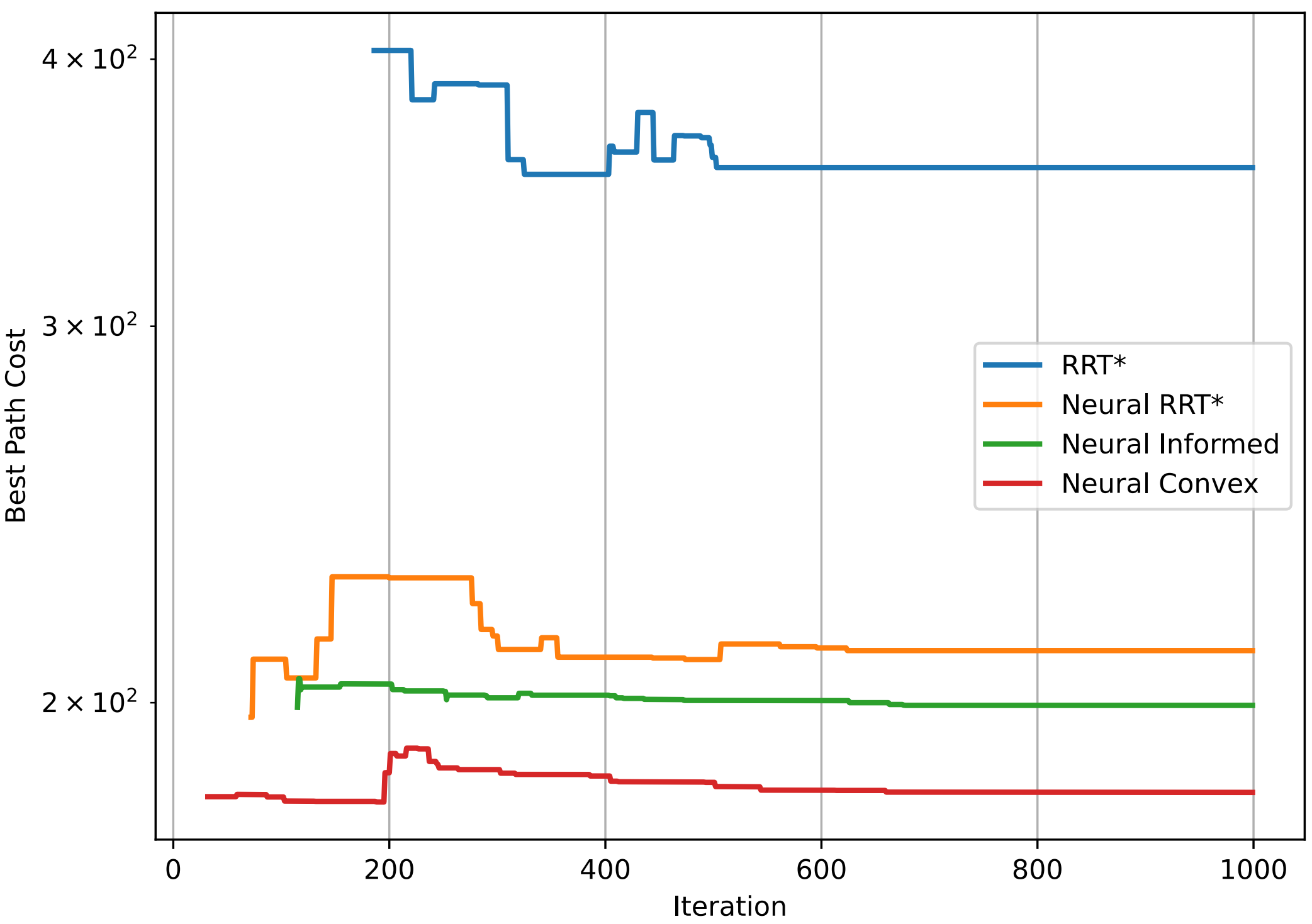

*Convergence curves in dense environments.*

## 4.5 Robustness to Prediction Noise

To assess the dependency on neural guidance, robustness was evaluated under Gaussian shift noise ($\sigma = 2{,}4$) and random deletion of predicted convex corners (30% and 60%). Performance degradation was measured relative to the clean baseline for each map category (Table [tab:robustness_all]).

In sparse and medium environments, path length variations remained below 0.7%, and success rates stayed at 100%, indicating stable performance even under significant perturbations. In hard environments, degradation was more noticeable but remained controlled, with a maximum path length increase of 1.37% and a success rate reduction below 10%.

These results indicate that Convex-Neural RRT* does not excessively depend on neural predictions and maintains reliable performance even when guidance is partially corrupted.

Although the neural model is trained on procedurally generated maps with polygonal obstacles, its performance in environments with significantly different geometries, such as curved or highly irregular obstacles, may be affected. However, the proposed method does not rely solely on neural predictions. The integration of convex-corner extraction and controlled exploration enables the planner to maintain robustness even when predictions are imperfect. This hybrid design allows the method to recover from prediction inaccuracies and preserve feasible path generation. Extending the training dataset to include more diverse environments and evaluating performance on real-world maps remain important directions for future work.

| Map Type | Noise Type | Length (%) | Time (%) | Smooth (%) | Success (%) |
|---|---|---|---|---|---|
| Sparse | $\sigma = 2$ Shift | +0.11 | -5.29 | +14.45 | 0.00 |
| | $\sigma = 4$ Shift | +0.34 | +0.86 | +10.54 | 0.00 |
| | 30% Deletion | +0.05 | +15.29 | +3.73 | 0.00 |
| | 60% Deletion | +0.16 | -15.05 | +25.60 | 0.00 |
| Medium | $\sigma = 2$ Shift | +0.20 | -10.01 | -3.72 | 0.00 |
| | $\sigma = 4$ Shift | +0.65 | -2.21 | -1.15 | 0.00 |
| | 30% Deletion | +0.35 | -16.53 | +7.12 | 0.00 |
| | 60% Deletion | +0.25 | -19.60 | +0.32 | 0.00 |
| Hard | $\sigma = 2$ Shift | +0.49 | -26.17 | +2.8 | -4.16 |
| | $\sigma = 4$ Shift | +0.62 | -29.63 | +1.3 | -9.16 |
| | 30% Deletion | +0.21 | -10.50 | +4.80 | -0.84 |
| | 60% Deletion | +1.37 | -18.04 | +12.49 | -7.56 |

# 5 Conclusion

This paper introduced Convex-Neural RRT*, a structured learned-guided sampling-based planner designed to address the limitations of classical RRT* in cluttered environments. The method integrates the incremental refinement mechanism of RRT* with a U-Net prediction model to identify informative waypoint regions, from which convex candidate regions are extracted. These regions enable focused sampling and accelerated tree expansion while maintaining practical convergence behavior comparable to RRT*.

Extensive experiments across 18 maps and multiple difficulty levels, compared against RRT*, Neural RRT*, Neural Informed RRT*, and LTA*, show that Convex-Neural RRT* achieves consistent improvements in computational efficiency while maintaining competitive path quality and a high success rate. In many scenarios, planning time is substantially reduced relative to neural-guided and classical baselines. The results indicate that convex-guided sampling provides an effective balance between computational cost and solution quality, supporting its potential suitability for real-time robotic navigation and compute-constrained systems.

Future work will extend Convex-Neural RRT* to full 3D UAV planning, where complex obstacle geometries and altitude variations require enhanced geometric reasoning. The current neural guidance module is trained on synthetically generated polygonal environments; therefore, we aim to extend both training and validation to real-world occupancy maps with irregular and curved obstacle geometries in order to further assess generalization and practical applicability. We also aim to investigate adaptive learned models that update predictions online from onboard sensors, enabling responsiveness to dynamic environments and moving obstacles. Integration with real-time UAV control frameworks represents another promising direction.

# Acknowledgment

This work was supported by the research project "Modeling and Control of Aerial Manipulators" N°
C00L07UN310220230004

Hichem Cheriet received the Master’s degree in Computer Science from ENS Kouba University, Algeria, in 2017. He is currently pursuing the Ph.D. degree in data science, robotics, and intelligent systems at USTO University, Algeria. His research interests include autonomous path planning, optimal motion planning, sampling-based algorithms, learning-guided planning, and control. He has authored and co-authored several journal and conference papers in the field of intelligent robotic navigation.

Badra Khellat Kihel is an Associate Professor and researcher at Oran2 University. She holds a Habilitation to Direct Research from the University of Science and Technology of Oran, where she also earned her PhD in Computer Science, specializing in pattern recognition, optimization, and feature selection.

Her current research focuses primarily on dimensionality reduction, a critical area in machine learning and data science aimed at simplifying complex datasets while preserving essential information. She has authored numerous scientific papers published in both national and international peer-reviewed journals, contributing significantly to advances in data preprocessing and intelligent systems.

Samira Chouraqui is a Lecturer and Researcher in Computer Science at the University of Science and Technology of Oran Usto'MB. She received her PhD degree in applied computer science from Usto'MB University in 2010. Her research interests include artificial intelligence, metaheuristic optimization algorithms, control systems, and intelligent autonomous systems.

She focuses particularly on the optimization of PID controllers for unmanned aerial vehicles (UAVs), drone modeling, and simulation.

Bara J.Emran earned his Ph.D. in Mechanical Engineering from the University of British Columbia in 2019. He has worked with several leading robotics companies across a range of industries, contributing to the development and application of advanced robotic systems. In 2024, he joined the Department of Electrical Engineering at the American University of Sharjah (AUS). His research interests lie at the intersection of advanced control systems and machine learning, with a particular focus on developing computationally efficient, learning-based control algorithms.